\documentclass[10pt,journal,compsoc]{IEEEtran}

\usepackage{url}
\usepackage{ragged2e}
\usepackage{epsfig}
\usepackage{graphicx}
\usepackage{amsmath,amssymb} % define this before the line numbering.
\usepackage{subfigure}
\usepackage{algorithm}
\usepackage{algorithmic}
\usepackage{graphics}
\usepackage{threeparttable}
\usepackage{color}
\usepackage[normalem]{ulem}
\usepackage{multirow}
\usepackage{float}
\usepackage{amsfonts}
\usepackage{bm}
\usepackage{array}
\usepackage{colortbl}
\usepackage{pifont}
\usepackage{diagbox}
\usepackage{rotating}
\usepackage{booktabs}
\usepackage{overpic}
\usepackage{textcomp}
\usepackage{contour}
\usepackage{enumitem}
\usepackage{ulem}
\usepackage{makecell}

\ifCLASSOPTIONcompsoc
  \usepackage[nocompress]{cite}
\else
  \usepackage{cite}
\fi

\usepackage[colorlinks=true]{hyperref} 
\begin{document}
\title{Say No to Freeloader: Protecting Intellectual Property of Your Deep Model}
\author{Lianyu~Wang, 
        Meng~Wang,
        Huazhu~Fu,~\IEEEmembership{Senior Member,~IEEE},\\
        and
        Daoqiang~Zhang,~\IEEEmembership{Senior Member,~IEEE,}% <-this % stops a space
\IEEEcompsocitemizethanks{\IEEEcompsocthanksitem L.~Wang and M.~Wang contributed equally to this work. Corresponding author: H.~Fu (hzfu@ieee.org) and D.~Zhang (dqzhang@nuaa.edu.cn).
\IEEEcompsocthanksitem L.~Wang and D.~Zhang are with the Key
Laboratory of Brain-Machine Intelligence Technology, Ministry of Education, Nanjing 211106, China. 
\IEEEcompsocthanksitem M.~Wang and H.~Fu are with the Institute of High Performance Computing (IHPC), Agency for Science, Technology and Research (A*STAR), 138632, Singapore.}% <-this % stops a space
}

% \markboth{Journal of \LaTeX\ Class Files,~Vol.~14, No.~8, August~2015}%
% {Shell \MakeLowercase{\textit{et al.}}: Bare Advanced Demo of IEEEtran.cls for IEEE Computer Society Journals} 

\IEEEtitleabstractindextext{%
\begin{abstract}
  \justifying
Model intellectual property (IP) protection has attracted growing attention as science and technology advancements stem from human intellectual labor and computational expenses.
Ensuring IP safety for trainers and owners is of utmost importance, particularly in domains where ownership verification and applicability authorization are required.
A notable approach to safeguarding model IP involves proactively preventing the use of well-trained models of authorized domains from unauthorized domains.
In this paper, we introduce a novel Compact Un-transferable Pyramid Isolation Domain (CUPI-Domain) which serves as a barrier against illegal transfers from authorized to unauthorized domains.
Drawing inspiration from human transitive inference and learning abilities, the CUPI-Domain is designed to obstruct cross-domain transfers by emphasizing the distinctive style features of the authorized domain. This emphasis leads to failure in recognizing irrelevant private style features on unauthorized domains.
To this end, we propose novel CUPI-Domain generators, which select features from both authorized and CUPI-Domain as anchors. Then, we fuse the style features and semantic features of these anchors to generate labeled and style-rich CUPI-Domain. Additionally, we design external Domain-Information Memory Banks (DIMB) for storing and updating labeled pyramid features to obtain stable domain class features and domain class-wise style features.
Based on the proposed whole method, the novel style and discriminative loss functions are designed to effectively enhance the distinction in style and discriminative features between authorized and unauthorized domains, respectively.
Moreover, we provide two solutions for utilizing CUPI-Domain based on whether the unauthorized domain is known: target-specified CUPI-Domain and target-free CUPI-Domain. By conducting comprehensive experiments on various public datasets, we validate the effectiveness of our proposed CUPI-Domain approach with different backbone models. The results highlight that our method offers an efficient model intellectual property protection solution.
\end{abstract}
 
}

\maketitle
 
\IEEEdisplaynontitleabstractindextext
\IEEEpeerreviewmaketitle

\ifCLASSOPTIONcompsoc
\IEEEraisesectionheading{\section{Introduction}\label{sec:introduction}}
\else
\section{Introduction}
\label{sec:introduction}
\fi

\IEEEPARstart{D}{eep} Neural networks have made significant strides in diverse areas of machine learning. However, recent achievements heavily relied on abundant and high-quality data~\cite{high-quality1, high-quality2}, dedicated training resources~\cite{training-resources1, training-resources2}, and meticulous manual fine-tuning~\cite{fine-tuning1, fine-tuning2}. Acquiring a proficiently trained deep model demands substantial time and effort in practice. While some DNN models, such as VGG~\cite{VGG}, Inception-v3~\cite{Inception}, and SWIN~\cite{SWIN}, are publicly available for non-commercial use, many owners of models used in commercial applications prefer to keep their trained DNN models private. This is often due to business considerations and concerns related to privacy and security, especially in critical applications like autonomous driving, face recognition, and intrusion detection~\cite{IPcm}.

Unfortunately, once these models are sold, they can be easily copied and redistributed, infringing on the interests of the developers and potentially increasing security risks in safety-critical applications, causing incalculable damage.
For instance, in commercial Machine Learning as a Service (MLaaS) platforms~\cite{mlaas}, well-trained deep models have high business value. They should be treated as the intellectual property (IP) of their creators/owners. When these models are uploaded to public platforms or deployed as remote services on the cloud, malicious users may steal them for financial gain. One of the most concerning threats is "Will releasing the model make it easy for the main competitor to copy this new feature and hurt owner differentiation in the market?". Therefore, it is crucial to regard and protect these models as valuable scientific and technological intellectual property (IP)~\cite{IP1, IP2, NTL}. 

\begin{figure}[!t]
  \centering
   \includegraphics[width=1\linewidth,trim=250 40 100 70,clip]{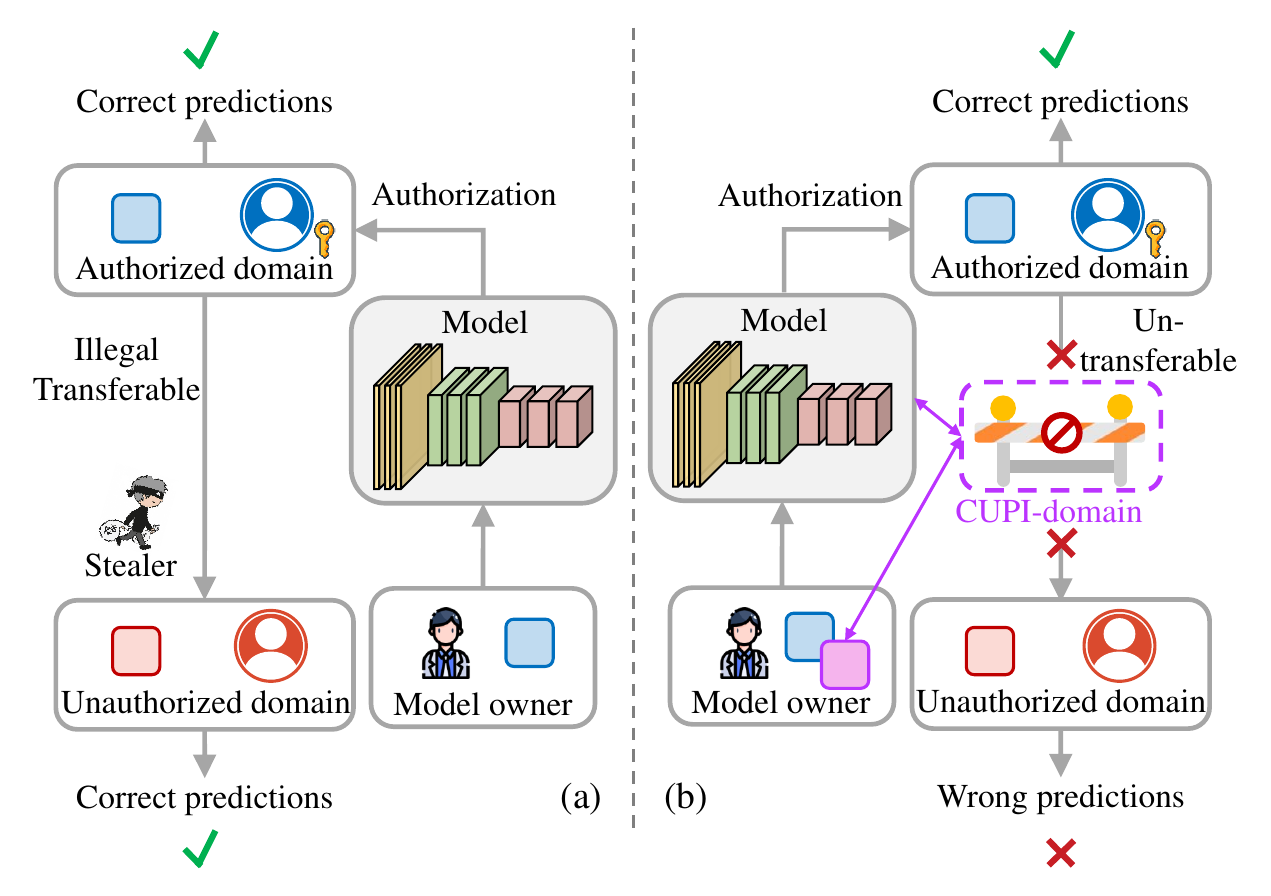}
   \caption{Model IP protection with our proposed CUPI-Domain. (a): In traditional supervised learning (SL), owners train deep models on the authorized domain (blue square), and then authorized users can gain correct predictions on the authorized domain. However, freeloaders can illicitly steal these models and deploy them on unauthorized domains (red squares) to obtain usable results for profit.
   (b): Our method constructs a CUPI-Domain between the authorized and unauthorized domains, which could block the illegal transferring and lead to a wrong prediction for unauthorized domains.}
   \label{figure1}
   \vskip -10pt
\end{figure}

Traditionally, model owners design, train, deliver, and deploy efficient deep models based on the authorized domain (represented by blue squares in Fig.~\ref{figure1}). They grant specific users the right to use their well-trained models, enabling them to obtain correct predictions on the authorized domain, as depicted in Fig.~\ref{figure1}~(a). However, since these models are trained with the overall features of the authorized domain, they may inadvertently cover unauthorized domains as well. Here, the unauthorized domain refers to a domain that shares the same task but exhibits different features compared to the authorized domain. This difference is primarily attributed to variations in the device, which in real cases is associated with the model owner rather than the user. This coverage has led to the emergence of freelancers who illegally steal well-trained deep models and deploy them on unauthorized domains (indicated by red squares in Fig.~\ref{figure1}). They process the models using techniques such as domain adaptation~\cite{DA1, DA2} or domain generation~\cite{DG1, DG2} to obtain correct predictions and claim these models as their own products for profit.
Such behavior infringes on the interests of model developers, and may even increase the potential security risks on safety-critical applications, causing incalculable damage and hindering the long-term development of deep model~\cite{long-term}. 
% For example, in the commercial Machine Learning as a Service (MLaaS) platforms~\cite{mlaas}, the well-trained deep models have high business value and should be considered as the IPs of the model creators/owners.
% One of the most concerning threats is "Will releasing the model make it easy for the main competitor to copy this new feature and hurt owner differentiation in the market?". 
Thus, protecting the rights of model owners and ensuring the safety of model deployment is essential.

Comprehensive model IP protection for deep models includes both ownership verification and applicability authorization~\cite{IP1, IP2, NTL, IP4}. 
Ownership verification refers to "Who can use the model". Model owners can grant usage permission to specific users, and any other users will be infringing on the owner's IP rights when employing the model~\cite{IP1, IP2}. However, it is important to acknowledge that there is no guarantee that the trained model will remain secure and not be leaked. There are two potential risks to consider: 1) Cyber hackers may attempt to steal the model from the cloud storage or other platforms where it is stored. 2) Specific users who have been granted permission may transfer the model to unauthorized or illegal users, which undermines the control and ownership of the model, potentially leading to misuse. These risks highlight the importance of implementing ownership verification mechanisms to protect against unauthorized access, sharing, and misuse of trained models. Common ownership verification mechanisms include embedding watermarks~\cite{watermarks1, watermarks2}, utilizing model fingerprints~\cite{fingerprints}, or employing predefined triggers~\cite{triggers}.
However, it is essential to recognize that these methods can be vulnerable to various techniques such as fine-tuning~\cite{F/RTAL}, classifier retraining~\cite{F/RTAL}, elastic weight consolidation algorithms~\cite{EWC/AU}, and watermark overwriting. These vulnerabilities can potentially weaken the effectiveness of the model's IP protection measures. Therefore, it is crucial to explore and develop more robust ownership verification techniques.

Applicability authorization refers to "Where can use the model". The model owner trains the deep model on the authorized domain and grants specific users the right to utilize the model within that domain, ensuring correct predictions~\cite{NTL, IP4}. However, in practice, legal users and illegal users may employ techniques such as domain adaptive~\cite{DA1, DA2}, domain generalization~\cite{DG1, DG2}, etc., to process the model and utilize it on other domains with similar tasks~\cite{NTL}, thereby achieving usable performance. This kind of infringement is not only easier to carry out but also more common and often concealed. This poses a significant challenge for maintaining control over the model's usage and protecting the IP of the model owner.
Therefore, it is necessary to develop a more robust applicability authorization mechanism that restricts the model's performance to the tasks specified by the owner, ensuring that the model's performance on unauthorized domains is not superior to that of an untrained model. This will discourage unauthorized users from attempting to steal the model.
To achieve this, a Non-Transferable Learning (NTL) method is proposed~\cite{NTL}, which uses an estimator with a characteristic kernel from Reproducing Kernel Hilbert Spaces to approximate and increase the maximum mean difference between two distributions on finite samples.
However, the authors only considered using limited samples to increase the mean distribution difference of features between domains and ignored outliers. The convergence region of NTL~\cite{NTL} is not tight enough.
Moreover, the calculation of the maximum mean difference is class-independent, which reduces the model's feature recognition ability in the authorized domain to a certain extent.

To address these challenges, we propose a novel Compact Un-transferable Pyramid Isolation Domain (CUPI-Domain) which serves as a barrier against illegal transfers from authorized to unauthorized domains, as depicted in Fig~\ref{figure1}~(b).
Our proposed CUPI-Domain considers the overall features of each domain, comprising two components: shared features and private features.
Shared features refer to semantic features, while private features include stylistic cues such as texture, saturation, perspective, brightness, background, etc. 
Initialized with the unauthorized domain, our proposed CUPI-Domain particularly emphasize the private features of the authorized domain, strategically fusing them with its semantic features, constructing richer style features for adaptively style augmentation.
By intentionally diminishing the recognition capability of the CUPI-Domain, we can implicitly impede illegal transfers to unauthorized domains that possess irrelevant private style features, thereby leading to wrong predictions. 
Additionally, we design Domain-Information Memory Banks (DIMB) for storing and updating labeled pyramid features to obtain stable domain class features and domain class-wise style features.
The novel style loss function is designed to effectively enhance the style difference between the authorized and unauthorized domains, whereas the discriminative loss functions are tailored to optimize the discriminative difference between the authorized/unauthorized domain and the CUPI-Domain.
Moreover, we further develop two distinct solutions for leveraging the CUPI-Domain, depending on whether the unauthorized domain is known or not: target-specified CUPI-Domain and target-free CUPI-Domain.

In general, we highlight our five-fold contributions:
\begin{itemize}
\item{\textbf{We propose a novel IP protection approach, named CUPI-Domain,} to block illegal transfers from authorized to unauthorized domains. \textbf{Meanwhile, CUPI-Domain generators are designed} to generate labeled and style-rich CUPI-Domain by emphasizing the private features of the authorized domain, implicitly leading to failure in recognizing irrelevant private style features on unauthorized domains.}
\item{\textbf{We introduce externel DIMB} to store specified features to obtain stable domain class features and domain class-wise style features for the subsequent computation.}
\item{\textbf{Style and discriminative loss functions are designed} to further improve inter-domain differences while maintaining semantic consistency between the unauthorized domain and CUPI-Domain.}
\item{\textbf{Two distinct solutions are developed for utilizing CUPI-Domain} based on whether the unauthorized domain is known or not: target-specified CUPI-Domain and target-free CUPI-Domain.}
\item{\textbf{Finally, comprehensive experimental results on several public datasets demonstrate that CUPI-Domain effectively reduces the recognition ability on unauthorized domains while maintaining strong recognition on authorized domains.} As a plug-and-play module, our CUPI-Domain can be easily implemented within different backbones and provide efficient solutions.\footnote[1]{https://github.com/LyWang12/CUPI-Domain.}}
\end{itemize}

This paper is based on and extends our previous CVPR2023 version~\cite{CVPR} in the following aspects.
1) We have implemented more standardized generators known as the \textbf{CUPI-Domain generators}. This generator effectively eliminates the style features of the input CUPI-Domain and subsequently integrates them with the style features of the authorized domain, resulting in enhanced similarity between the two. 
2) We also designed external \textbf{DIMB}, which store and update labeled pyramid features to obtain stable domain class features and domain class-wise style features for the subsequent computation. 
3) We designed novel \textbf{style loss function and discriminative loss functions} to improve the style difference and discriminative difference between authorized and unauthorized domains, while maintain the semantic consistency between the unauthorized domain and CUPI-Domain. 
4) We conducted additional experiments on the \textbf{Office-home-65 (65 categories, 4 domains)~\cite{Home}, and DomainNet (345 categories, 6 domains)~\cite{DomainNet}} to demonstrate the continued effectiveness of our method as the data complexity increases. 
5) We have invested substantial efforts in enhancing the presentation and organization of our paper, specifically focusing on the introduction, framework, key results, and discussion. We have provided more comprehensive explanations in the introduction section to offer a clearer understanding of the research context. Beside, we have introduced several new sections to elaborate on our novel framework, encompassing the method formulation, corresponding technical components, and loss functions. Additionally, we have diligently rewritten multiple sections to improve readability and provide more detailed explanations, particularly in the sections addressing quantitative comparisons, ablation experiments and discussion. These revisions aim to enhance the clarity and coherence of our paper, offering a more comprehensive and insightful reading experience.

\section{Related Work}
\subsection{Model IP Protection}
Currently, there are two primary categories of methods for protecting model IP: ownership verification and applicability authorization. In terms of ownership verification, watermark embedding~\cite{embedding1} is a widely used classic method. Kuribayashi~\textit{et al.}~\cite{embedding2} proposed a quantifiable watermark embedding method that aims to minimize the variations caused by embedding watermarks. Adi~\textit{et al.}~\cite{embedding3} introduced a black-box tracking mechanism for ownership verification. Zhang~\textit{et al.}~\cite{embedding4} proposed a model watermarking framework that utilizes spatial invisible watermarking to protect image processing models. 
However, it has been observed that watermark embedding approaches are vulnerable to certain watermark removal methods. Our experiments employ a simple watermark embedding in the model to verify ownership by triggering misclassification. Through comprehensive experimental results, we demonstrate that the proposed CUPI-Domain exhibits resistance against common methods of watermark removal.

Applicability authorization is an extension of usage authorization, where model owners typically employ a preset private key to encrypt the entire or partial network. This encryption ensures that only authorized users can obtain the private key and subsequently use the model. Various advanced methods have been developed for usage authorization. For instance, Alam~\textit{et al.}~\cite{usage1} introduced an explicit locking mechanism that utilizes S-Boxes with strong cryptographic properties to secure each training parameter of a lightweight deep neural network. Unauthorized access without knowledge of the legitimate private key can significantly degrade the model's accuracy. Song~\textit{et al.}~\cite{usage2} analyzed and calculated the critical weight of deep neural network models and reduced time costs by encrypting these critical weights to protect against unauthorized use. In terms of applicability authorization, Wang~\textit{et al.}~\cite{NTL} proposed a data-based approach called Non-Transferable Learning (NTL), which aims to preserve model performance on authorized data while intentionally degrading performance in other data domains~. This method allows the model to act well within authorized domains while exhibiting reduced effectiveness when applied to unauthorized domains.

In contrast to these methods, we propose a novel approach that involves constructing a new class-dependent CUPI-Domain with infinite samples. This CUPI-Domain is designed to have features that are more similar to the authorized domain. By deliberately reducing the model's performance on both the CUPI-Domain and the target domain, we can achieve a tighter generalization bound for the well-trained model, thereby constraining the model's performance within the authorized domain.

\subsection{Domain Transferring}
In recent years, deep learning models have achieved revolutionary success in diverse areas of machine learning. To inherit the advantages of deep models, domain adaptive~\cite{DA1, DA2} and domain generalization~\cite{DG1, DG2} have been proposed to rapidly improve the performance of well-trained deep models on target datasets. 

\begin{figure*}[!t]
  \centering
   \includegraphics[width=0.8\linewidth,trim=0 90 0 90,clip]{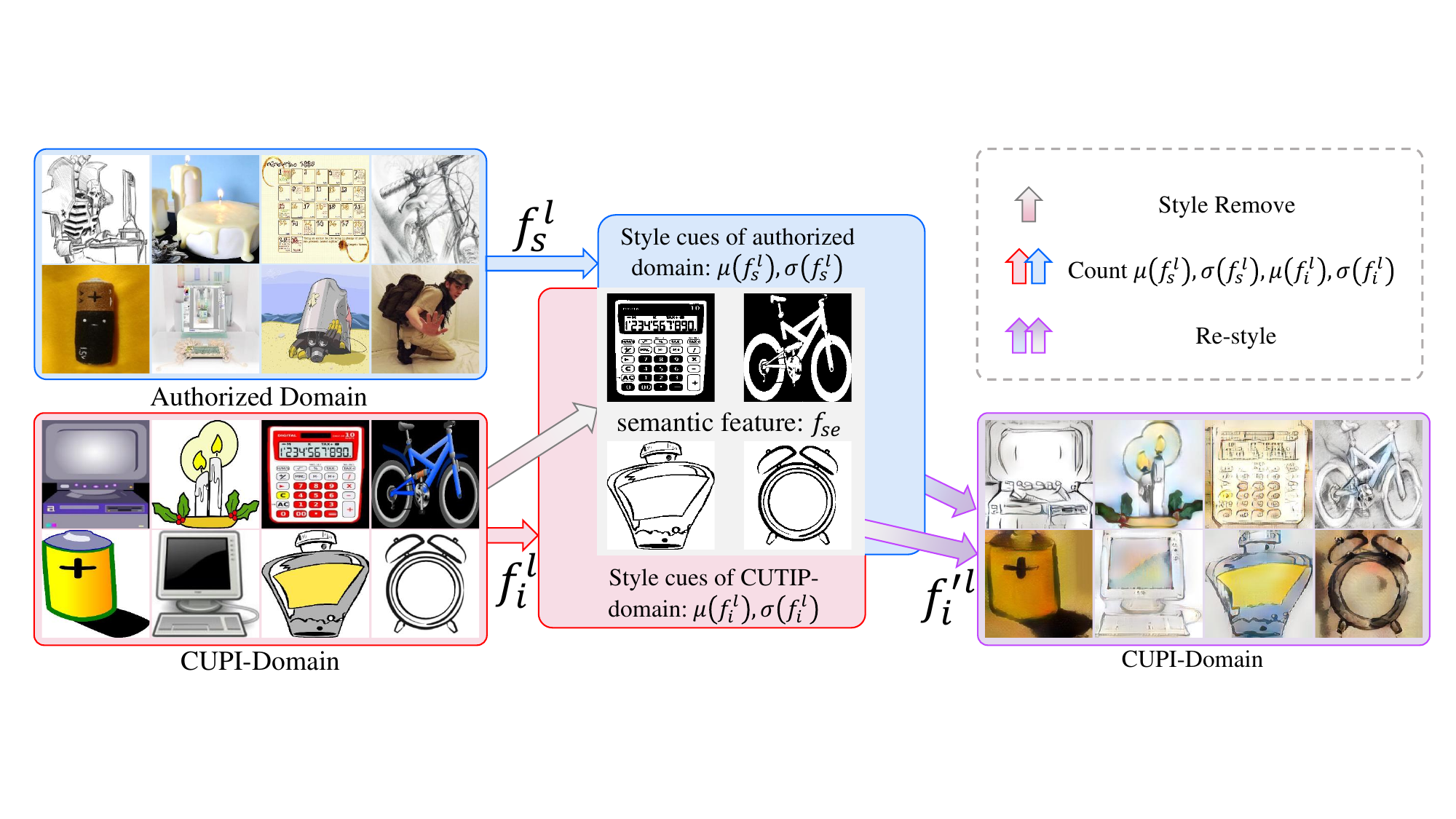}
   \caption{Illustration of our proposed CUPI-Domain generator. The output feature of the authorized domain \(f_s^l\) and the CUPI-Domain \(f_i^l\) in the \(l\)-th feature extractor block are fed into the CUPI-Domain generator, and then the style features of \(f_s^l\) are combined with the semantic features of \(f_i^l\) to update the CUPI-Domain. This update process ensures that the private style features of the updated \(f{'}_i^l\) are more similar to the features of the authorized domain \(f_s^l\) while preserving their original semantic features.}
   \label{figure2}
   \vskip -10pt
\end{figure*}

Domain adaptive~\cite{DA1, DA2} involves transferring a model from a labeled source domain to an unlabeled but relevant target domain, with the target domain's data being accessible during the training process~\cite{DA3}. Sun~\textit{et al.}~\cite{DUDA1} and Zhuo~\textit{et al.}~\cite{DUDA2} minimized inter-domain differences by aligning the second-order statistics of the source and target distributions. Saito~\textit{et al.}~\cite{AUDA} adopted a new adversarial paradigm, where the adversarial way occurs between the feature extractor and the classifier, rather than between the feature extractor and the domain discriminator.

Domain generalization~\cite{DG1, DG2} differs from domain adaptive in that the target domain is inaccessible during training phase\cite{DG3, DG4}. Tobin~\textit{et al.}~\cite{DG3} used domain randomization to generate more training data from simulated environments for generalization in real environments. Prakash~\textit{et al.}~\cite{DG4} further considered the structure of the scene when randomly placing objects for data generation, enabling the neural network to learn how to utilize context when detecting objects.
Recently, several methods have been proposed and effectively applied to cross-domain applications. These methods aim to find an intermediate state that lies between the source domain and the target domain. By emphasizing the similarity between domains, they enhance the transferability of the model~\cite{Med1,Med2,Med3,Med4,Med5}. 

In contrast, the objective of our proposed CUPI-Domain is to explore an intermediate state that accentuates the distinctions between the authorized and unauthorized domains, thereby limiting the transferability of the model and ensuring the protection of the model owners' intellectual property with respect to their scientific and technological achievements.

\section{Methodology}
We first provide a comprehensive explanation of our proposed CUPI-Domain in Section 3.1, which serves as a barrier to restrict the model's performance within the authorized domain and diminish its feature recognition capabilities on unauthorized domains. Subsequently, we introduce external DIMB for storing and updating labeled pyramid features in Section 3.2. Additionally, we present two distinct solutions based on whether the unauthorized target domain is known or not: the target-specified CUPI-Domain and the target-free CUPI-Domain in section 3.3, these solutions are designed to safeguard the IP of the model.

\subsection{Compact Un-transferable Pyramid Isolation Domain (CUPI-Domain)}
In the deep neural network model, the overall features extracted by the feature extractor include two abstract components, \textit{i.e.}, shared features, and private features~\cite{twofeature1, twofeature2}. Shared features refer to semantic features which reflect the structural information of samples and play a leading role in sample recognition~\cite{style3}. Private features refer to a collection of subtle, weak semantically related cues in the feature, such as lighting conditions, texture, hue, color saturation, and background~\cite{style1}.
Samples from different domains with the same task usually exhibit consistent semantic information while significant stylistic variations. 

Most previous works on domain adaptive and domain generalization have primarily focused on enhancing feature transferability between domains. This has involved strengthening the model's emphasis on shared features while suppressing private style features that may appear as disturbances. However, to ensure the protection of the model's IP, this paper aims to restrict the model's feature recognition capability by accentuating the private style features of the authorized domain through style transfer techniques. This emphasis leads to failure in recognizing irrelevant private style features on unauthorized domains.

\begin{figure*}[!t]
\centering
\includegraphics[width=0.8\linewidth,trim=60 20 70 60,clip]{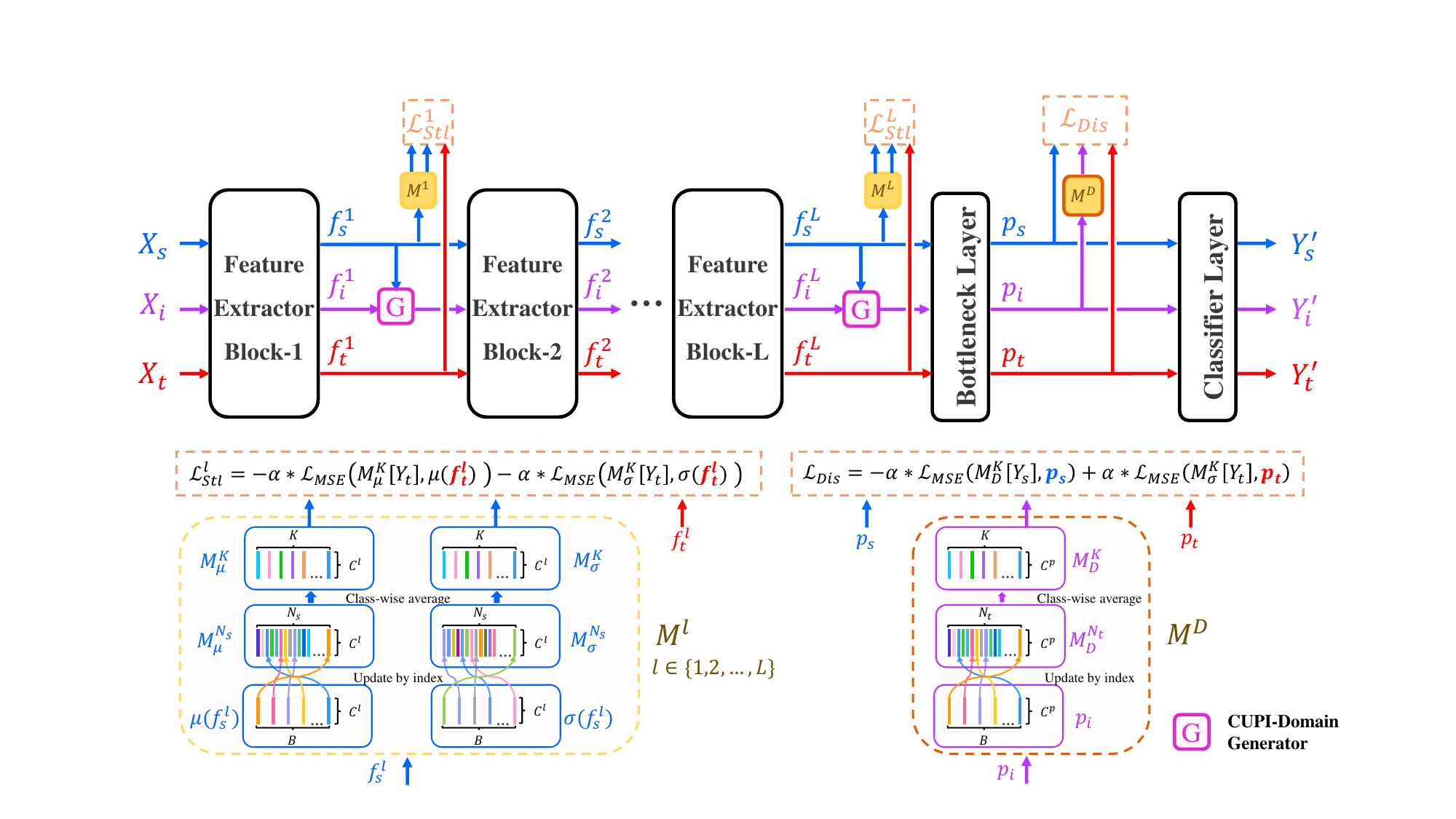}
\caption{The illustration of our proposed CUPI-Domain, including the feature extractor blocks, CUPI-Domain generators, the external DIMB, and style/distinctive loss functions. The samples from the authorized domain, CUPI-Domain, and unauthorized domain are fed into the feature extractor in parallel, denoted by blue, purple, and red, respectively.
The CUPI-Domain generators are deployed after each feature extractor block, while the external DIMB are designed to store specified labeled features to obtain stable domain class features and domain class-wise style features for subsequent computation.}
\label{figure3}
\end{figure*}

Style transfer studies have conjectured that styles exhibit homogeneity and consist of repeated structural motifs. In this context, two images are considered to have similar styles if the features extracted by a trained classifier exhibit shared statistics~\cite{style1, style2}, such as first- and second-order statistics, which are commonly used due to their computational efficiency. First- and second-order statistics refer to the mean and variance of the extracted features, which are called style features.
Semantic features capture pure structural information without incorporating style cues. Following Dumoulin~\textit{et al.}~\cite{style3}, the semantic features \(f_{se}\) of the extracted feature \(f\) can be obtained by removing style features as: 
\begin{equation}
\label{eq1}
f_{se} = \frac{{f - {\mu(f)}}}{{{\sigma(f)}}},
\end{equation} 
where \(\mu(f)\) and \(\sigma(f)\) denote the mean and variance of \(f\). Furthermore, style can be re-assign by \(f \cdot \gamma  + \beta\), where \(\gamma\) and \(\beta\) are learned parameters. Afterward, Huang~\textit{et al.}\cite{style1} further explored adapting \(f\) to an arbitrarily given style by using style features of another extracted feature instead of learned parameters.

Building upon these concepts, we present a novel CUPI-Domain that combines style features from the authorized domain with semantic features from the CUPI-Domain. We achieve this by initializing the CUPI-Domain with the unauthorized domain and then randomly extracting style features from the authorized domain, integrating them with the semantic features of the CUPI-Domain. This process results in the private style features of the CUPI-Domain becoming more similar to those of the authorized domain. By employing this strategy, we effectively diminish the model's feature recognition capabilities on both the CUPI-Domain and the unauthorized domain. Consequently, we implicitly obstruct the pathway between the authorized and unauthorized domains, thereby confining the model's performance exclusively to the authorized domain.

Our CUPI-Domain is generated by the CUPI-Domain generator, which is a lightweight, and plug-and-play module, as depicted in Fig.~\ref{figure2}. \(f_s^l\) and \(f_i^l\) represent the deep features of the \(l\)-th feature extractor block in the authorized domain and the CUPI-Domain, respectively. To begin, \(f_s^l\) and \(f_i^l\) are concurrently fed into the CUPI-Domain generator. Subsequently, the mean \(\mu(f_s^l)\) and variance \(\sigma(f_s^l)\) of \(f_s^l\) are calculated along the channel dimension to represent private style features of the authorized domain, followed by a 1×1 convolution layer \(Conv\). Next, removing style features of \(f_i^l\). Finally, the \(\mu(f_s^l)\) and variance \(\sigma(f_s^l)\) of \(f_s^l\) are multiplied and added channel-wisely as:

\begin{equation}
\label{eq2}
f{'}_i^{l} = \frac{{f_i^l - {\mu(f_i^l)}}}{{{\sigma(f_i^l)}}} \bigotimes Conv(\sigma (f_s^l)) \bigoplus Conv(\mu(f_s^l)).
\end{equation} 

Subsequently, \(f_s^l\), \(f{'}_i^{l}\), and \(f_t^l\) are simultaneously fed into the  \(l+1\)-th feature extractor block in parallel.
Through continuous feature fusion, as shown in Fig.~\ref{figure2}, CUPI-Domain generator can build style-rich latent feature space for each class and then construct a labeled CUPI-Domain containing similar private styles to the authorized domain.

\subsection{Domain-Information Memory Banks (DIMB)}
In our method, we employ external DIMB comprising multiple memory cells to store and update domain-dependent pyramid features via sample index. This enables us to obtain stable domain class features and domain class-wise style features throughout the training process, as depicted at the bottom of Fig.~\ref{figure3}.
In this section, we provide a comprehensive description of the workflow associated with our proposed DIMB.

\subsubsection{Initialization Strategy of DIMB}
Before training, the pre-trained feature extractor with frozen model parameters is utilized to initialize the DIMB as follows:

\begin{itemize}
\item{Since the output features of the \(l\)-th feature extractor \(f_s^l\in {\mathbb{R}^{B \times N^l \times {C^l}}}\) in the authorized domain contain rich semantic information with style clues, and feature vector \(p_i\in {\mathbb{R}^{B \times {C^p}}}\) before the classifier layer contains refined class-discriminative information in the CUPI-Domain, we store \(f_s^l\) and \(p_i\) in the memory cell \(M^l\) and \(M^D\) respectively. Where \(B\) denotes the batch size of the input, \(N^l\) represents the output size of the \(l\)-th feature extractor, \(C^l\) and \(C^p\) denotes the feature dimension of the \(l\)-th feature extractor and bottleneck layer, respectively.}
\item{Calculate the channel-wise mean $\mu \left({f_s^l} \right) \in {\mathbb{R}^{B \times {C^l}}}$ and variance $\sigma \left({f_s^l} \right) \in {\mathbb{R}^{B \times {C^l}}}$  of \(f_s^l\), and update the corresponding \(B\) features in $M_\mu ^{{N_s}} \in {\mathbb{R}^{{N_s} \times {C^l}}}$ and $M_\sigma ^{{N_s}} \in {\mathbb{R}^{{N_s} \times {C^l}}}$ according to the sample index, meanwhile, do the same with \(p_i\) to update $M_D^{N_t} \in {\mathbb{R}^{{N_t} \times {C^p}}}$. Where \(N_s\) and \(N_t\) denote the number of samples in the authorized domain and the unauthorized domain respectively.}
\item{Repeat the above steps until all samples in the authorized domain have been traversed. Calculate the class centers of all features in $\left\{ {M_\mu ^{{N_s}},M_\sigma ^{{N_s}},M_D^{{N_t}}}\right\}$ according to their class labels and store the result in $\left\{ {M_\mu ^K \in {\mathbb{R}^{K \times {C^l}}},M_\sigma ^K \in {\mathbb{R}^{K \times {C^l}}}, M_D^K \in {\mathbb{R}^{K \times {C^p}}}} \right\}$}, respective, where \(K\) denote the class number. Finally, unfreeze the model parameters.
\end{itemize}

\subsubsection{Update Strategy of DIMB}
During training, we directly discard those obsolete features \(f_s^l\) and \(p_i\) according to the index of the samples and replace them with the latest features, then calculate and update the corresponding features in memory cell $ M^l = \left\{M_\mu ^{{N_s}}, M_\sigma ^{{N_s}}, M_\mu ^{K}, M_\sigma ^{K}\right\} (l \in {1,2,…, L})$  and  $ M^D = \left\{M_D^{{N_t}}, M_{D}^K \right\}$ as in the initialization strategy. Among them, \(M_D^K\) stores \(K\) class features, while \(M_\mu ^K\) and \(M_\sigma ^K\) store style features of \(K\) classes, which are used for downstream computation. Furthermore, the proposed DIMB is only an external repository and does not participate in the back-propagation calculation of the network.

\subsection{Model IP Protection with CUPI-Domain}
\subsubsection{Target-Specified CUPI-Domain}
We first introduce our solutions for model IP protection with a given unauthorized target domain, termed target-specified CUPI-Domain.

Fig.~\ref{figure3} provides an overview of the entire framework trained using our proposed CUPI-Domain. The framework comprises  \(L\) feature extractor blocks, \(L\) CUPI-Domain generators, \(L+1\) external DIMB, a bottleneck layer, and a classifier layer. The data from the authorized domain, CUPI-Domain, and unauthorized domain are denoted as \(X_s\), \(X_i\) (Initialized by \(X_t\)), and \(X_t\), respectively. During the training process,  \(X_s\), \(X_i\), and \(X_t\) are simultaneously fed into the feature extractor blocks, followed by a CUPI-Domain generator. The classifier at the end of the network is responsible for predicting the category of the sample, and the corresponding prediction results are represented as \(Y_s'\), \(Y_i'\), and \(Y_t'\), respectively. 

Based on the designed framework and DIMB, we proposed a novel joint loss function that incorporates the classification loss \({{\cal L}_{cls}}\), style loss \({{\cal L}_{Stl}}\) and discriminative loss \({{\cal L}_{Dis}}\) to enhance the distinction between authorized and unauthorized domains, while maintaining the classification performance of authorized domain. The overall loss function \(\cal L\) is as follows:
\begin{equation}
\label{eq3}
{{\cal L}} = {{\cal L}_{cls}} + {{\cal L}_{Stl}} + {{\cal L}_{Dis}}.
\end{equation}

The classification loss function \({{\cal L}_{cls}}\) is devised to enhance the class recognition capability within the authorized domain while reducing the class recognition ability of the CUPI-Domain and unauthorized domain, which consists of the respective domain loss functions of the authorized domain \({{\cal L}_s}\), CUPI-Domain \({{\cal L}_i}\) and unauthorized domain \({{\cal L}_t}\), as:
\begin{equation}
\label{eq4}
{{\cal L}_{cls}} = {\cal L}_s - {\cal L}_i - {\cal L}_t.
\end{equation}

For all three domains, the network can be trained using the Kullback-Leibler divergence loss function, which serves as a measure of dissimilarity between the predicted distribution and the target distribution:
\begin{align}
\label{eq5}
{{\cal L}_s} = \alpha * KL(Y_s'||Y_s), \\
\label{eq6}
{{\cal L}_i} = \alpha * KL(Y_i'||Y_i), \\
\label{eq7}
{{\cal L}_t} = \alpha * KL(Y_t'||Y_t),
\end{align}
where \(KL(\cdot)\) stands for Kullback–Leibler divergence, \(Y_s\), \(Y_i\) and \(Y_t\) denote the label of the authorized domain, CUPI-Domain and the unauthorized domain, respectively. The rising factor \(\alpha = {\left( {\frac{{epoch \_ number}}{{total \_ epoch}}} \right)^{0.9}}\) is used to speed up the convergence of the model on the authorized domain in the early stage of training. Additionally, to mitigate the impact of infinite amplification, an upper bound is set to limit the negative loss function, as suggested by Wang \textit{et al.}~\cite{NTL}.

The style loss function \({{\cal L}_{Stl}}\) is designed based on pyramid DIMB to enhance the distinction in style between authorized and other domains. It achieves this by incorporating stable multi-scale authorized domain style information, and real-time multi-scale style information of CUPI-Domain and unauthorized domain:
\begin{equation}
\label{eq9}
\begin{aligned}
& {{\cal L}_{Stl}} = - \alpha * \sum_{l=1}^L{\cal L}_{MSE}(M_\mu^{(l)K}[Y_t], \mu(f_t^l)) \\
& \quad \quad \quad - \alpha * \sum_{l=1}^L{\cal L}_{MSE}(M_\sigma^{(l)K}[Y_t], \sigma(f_t^l)).
\end{aligned}
\end{equation}  
Where \({\cal L}_{MSE}\) denotes the mean squared error.
The discriminative loss \({{\cal L}_{Dis}}\) is used to maintain the semantic consistency between the restyled CUPI-Domain and the unauthorized domain while enhancing the difference between CUPI-Domain and the authorized domain, which is defined as follows:
\begin{equation}
\label{eq10}
{{\cal L}_{Dis}} = - \alpha * {\cal L}_{MSE}(M_D^K[Y_s], p_s) + \alpha * {\cal L}_{MSE}(M_D^K[Y_t], p_t).
\end{equation} 

We summarize the strategy of our proposed target-specified CUPI-Domain in Supplementary Algorithm~1.

\subsubsection{Target-Free CUPI-Domain}
To tackle scenarios where the unauthorized domain is unknown, we introduce the Target-free CUPI-Domain. In such cases, it is not feasible to directly incorporate the unauthorized domain and CUPI-Domain into the model training process. To address this challenge, a synthesized unauthorized domain can be employed instead. For instance, Wang~\textit{et al.}~\cite{NTL} proposed a GAN-based approach that freezes parameters to generate synthesized samples in various directions as a substitute for the unauthorized domain training set. Although their method is capable of producing high-quality synthesized samples, the generator's direction is predetermined, leading to certain limitations. Additionally, Huang~\textit{et al.}~\cite{style1} developed an adaptive instance normalization (AdaIN) technique based on GAN, which can generate synthesized images with a specific style to complement a given content image.

To fully leverage the benefit, we introduce Gaussian noise to the AdaIN technique, enabling the generation of synthesized samples with random styles. Finally, we combine the synthesized samples obtained by the two aforementioned methods (\textit{e.g.}, GAN and AdaIN) to replace the unauthorized domain training set and initialize the CUPI-Domain. 

It is important to emphasize that our focus is on evaluating the effectiveness of the CUPI-Domain in preventing unauthorized feature transfer specifically within the context of synthesized images.
Our primary focus remains on reducing the feature recognition ability of the model on unauthorized domains while preserving it on the authorized domain. Any data synthesis method could be utilized in our work for generating the unauthorized domain.

The framework of the Target-free CUPI-Domain is consistent with the target-specified CUPI-Domain and the training process is detailed in Supplementary Algorithm~2.

During the testing phase, the model is evaluated on the authorized domain test set and other unknown domains with the same task. 

\section{Experiment}
\subsection{Datasets}
We evaluate our proposed CUPI-Domain on several popular domain adaption/generation benchmarks:
\begin{enumerate}
     \item \textbf{Digit datasets: MNIST~\cite{MT}, USPS~\cite{US}, SVHN~\cite{SN} and MNIST-M~\cite{MM}} are widely used digit datasets, each consisting of ten digits ranging from 0 to 9. These datasets include samples extracted from various scenes, providing a diverse range of digit images for evaluation and analysis.
     \item \textbf{CIFAR10~\cite{refCIFAR10} and STL10~\cite{refSTL10}} are both ten-class classification datasets commonly used in computer vision research. To ensure consistency between the datasets, we adopt the processing procedure outlined by French~\textit{et al.}~\cite{refSTLp}. This ensures that the classes in both datasets are aligned and can be directly compared during experimentation and evaluation.
     \item \textbf{VisDA-2017~\cite{VisDA}} is a Synthetic-to-Real dataset that consists of training (VisDA-T) and validation (VisDA-V) sets, each containing samples from 12 different categories.
     \item \textbf{Office-Home-65~\cite{Home}} consists of images from four distinct domains: Artistic, Clipart, Product, and RealWorld. Each domain contains 65 object categories, resulting in a total of 15,500 images in the dataset.
     \item \textbf{DomainNet~\cite{DomainNet}} consists of images from six domains, including clipart, infograph, painting, quickdraw, real and sketch, with a total of 345 scene-level categories. 
\end{enumerate}

\begin{table*}[!t]
\renewcommand\arraystretch{1.4}
  \centering
    \caption{The accuracy ($\%$) of target-specified CUPI-Domain on digit datasets. The vertical/horizontal axis denotes the authorized/unauthorized domain. In each task, the left of '\(\Rightarrow\)' shows the test accuracy of SL on the unauthorized domain, while the right side presents the accuracy of CUPI-Domain. 'Authorized/Unauthorized Drop' indicate the drop (relative drop) of CUPI-Domain relative to SL on authorized/unauthorized domains.}
  \resizebox{0.9\textwidth}{!}{
  \begin{tabular}{c|cccc|cc}
    \toprule
    Authorized/Unauthorized &  MNIST & USPS & SVHN & MNIST-M & Authorized Drop$\downarrow$ & Unauthorized Drop$\uparrow$ \\
    \midrule   
    MNIST~\cite{MT}   & 99.2 $\Rightarrow$  99.2 & 98.0 $\Rightarrow$ \ 6.5 & 38.2 $\Rightarrow$ \ 6.0 & 67.8 $\Rightarrow$ \ 8.5 & 0.00 (0.00\%) & 61.00 (88.37\%) \\ 
    USPS~\cite{US}    & 92.6 $\Rightarrow$ \ 9.1 & 99.7 $\Rightarrow$  99.7 & 25.5 $\Rightarrow$ \ 5.4 & 41.2 $\Rightarrow$ \ 8.1 & 0.00 (0.00\%) & 45.57 (83.11\%) \\ 
    SVHN~\cite{SN}    & 66.7 $\Rightarrow$ \ 9.5 & 70.5 $\Rightarrow$ \ 6.6 & 91.2 $\Rightarrow$  91.1 & 34.6 $\Rightarrow$ \ 7.8 & 0.07 (0.08\%) & 49.30 (84.62\%) \\ 
    MNIST-M~\cite{MM} & 98.4 $\Rightarrow$ \ 6.4 & 88.4 $\Rightarrow$ \ 7.4 & 46.3 $\Rightarrow$ \ 5.0 & 95.4 $\Rightarrow$  95.4 & 0.00 (0.00\%) & 71.43 (91.44\%) \\ 
    \bottomrule
  \end{tabular}}
  \label{ts_digit_detail}
\end{table*}

\begin{table*}[!t]
\renewcommand\arraystretch{1.4}
  \centering
    \caption{'Authorized/Unauthorized Drop' of target-specified CUPI-Domain, CUTI-Domain~\cite{CVPR} and NTL~\cite{NTL} on digit datasets. 
    % The result of CUTI-Domain~\cite{CVPR} and NTL~\cite{NTL} is calculated from the original paper. 
    The best performance is indicated by the numbers in bold. Statistical significance (p-value $<$ 0.05~\cite{P1,p2}) is denoted with: $^{\ast}$(CUPI-Domain vs. CUTI-Domain~\cite{CVPR}) and $^{\diamond}$(CUPI-Domain vs. NTL~\cite{NTL}).}
  \resizebox{0.9\textwidth}{!}{
  \begin{tabular}{c|ccc|ccc}
     \toprule
     \multirow{2}{*}{Domain / Drop} & \multicolumn{3}{c|}{Authorized Drop$\downarrow$} & \multicolumn{3}{c}{Unauthorized Drop$\uparrow$} \\
     \cline{2-7} 
     & CUPI-Domain & CUTI-Domain~\cite{CVPR} & NTL~\cite{NTL} & CUPI-Domain & CUTI-Domain~\cite{CVPR} & NTL~\cite{NTL} \\
    \midrule   
    MNIST~\cite{MT}   & \bf 0.00 (0.00\%) & 0.10 (0.10\%) & 1.00 (1.01\%) & \bf 61.00 (88.37\%) & \bf 61.00 (88.56\%) & 46.57 (75.60\%) \\ 
    USPS~\cite{US}    & \bf 0.00 (0.00\%) & 0.10 (0.10\%) & 1.00 (1.00\%) & \bf 45.57 (83.11\%) & 44.70 (80.72\%) & 38.67 (75.55\%) \\ 
    SVHN~\cite{SN}    & \bf 0.07 (0.08\%) & 0.30 (0.33\%) & 1.10 (1.23\%) & \bf 49.30 (84.62\%) & 48.33 (81.73\%) & 40.60 (77.25\%) \\ 
    MNIST-M~\cite{MM} & \bf 0.00 (0.00\%) & 0.00 (0.00\%) & 2.10 (2.30\%) & \bf 71.43 (91.44\%) & 69.73 (88.75\%) & 60.10 (76.95\%) \\ 
    \midrule
    Mean & \bf 0.02 (0.02\%)$^{\ast \diamond}$ & 0.13 (0.13\%) & 1.30 (1.39\%) & \bf 56.83 (86.89\%)$^{\ast \diamond}$ & 55.94 (84.94\%) & 46.48 (76.34\%) \\
    \bottomrule
  \end{tabular}}
  \label{ts_digit_compare}
\end{table*}

\begin{figure*}[!t]
  \centering
   \includegraphics[width=0.9\linewidth,trim=5 185 5 185,clip]{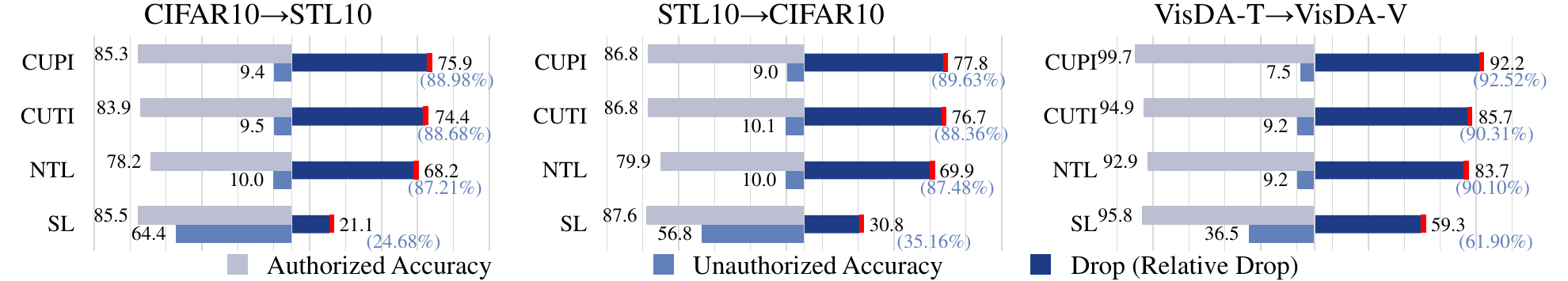}
   \caption{The accuracy ($\%$) of SL, target-specified NTL~\cite{NTL}, CUTI-Domain~\cite{CVPR}, and CUPI-Domain on CIFAR10\&STL10, and VisDA-2017. The title in each subgraph implies the authorized domain (left of '\(\rightarrow\)') and the unauthorized domain (right of '\(\rightarrow\)'). The bars with different colors represent the accuracy of each method in the authorized domain, the unauthorized domain, and the drop (relative drop) of the model performance, respectively.}
   \label{ts_csv}
\end{figure*}

\subsection{Implementation Details}
The implementation of our comprehensive experiments is based on the platform Pytorch and an NVIDIA GeForce RTX 3090 GPU with 24GB of memory. The Adam optimizer with an initial learning rate of 0.0001 is adopted to optimize the model. The batch size for each domain and the epoch number is set to 32 and 30, respectively.
To accommodate tasks of varying complexity, we employ different backbone architectures for various datasets followed by Wang~\textit{et al.}~\cite{NTL} for fair comparison. Specifically, we used VGG-11~\cite{VGG} for digit datasets, VGG-19~\cite{VGG} for CIFAR10 \& STL10 and VisDA-2017~\cite{VisDA}, and SWIN~\cite{SWIN} for Office-Home-65~\cite{Home} and DomainNet~\cite{DomainNet}. Pre-trained models are used for a fair comparison. We set \(L\) to 4 for SWIN and 5 for VGG to align with the number of feature extractor blocks in each architecture. Consistent with standard evaluation protocols, we utilize accuracy ($\%$) as the primary performance metric for each task.

\subsection{Result of Target-Specified CUPI-Domain}

\begin{table*}[!t]
\renewcommand\arraystretch{1.4}
  \centering
  \caption{The average results of ownership verification by SL and CUPI-Domain, CUTI-Domain~\cite{CVPR} and NTL~\cite{NTL}. 'Training Methods' shows SL and CUPI-Domain accuracy on the authorized domain without/with watermark. ‘Watermark Removal Approaches on CUPI’ represents CUPI-Domain accuracy after/before attacks by watermark removal methods: FTAL~\cite{F/RTAL}, RTAL~\cite{F/RTAL}, EWC~\cite{EWC/AU}, AU~\cite{EWC/AU}, and watermark overwriting followed by NTL~\cite{NTL}. 'Average Drop' represents the average decrease in test dataset accuracy with a watermark compared to without, after applying watermark removal methods. 
  %The result of CUTI-Domain~\cite{CVPR} and NTL~\cite{NTL} is obtained by reproducing its source code for a fair comparison. 
  The best performance is indicated by the numbers in bold.}
  \resizebox{0.9\textwidth}{!}{
  \begin{tabular}{c|cc|ccccc|ccc}
    \toprule
    \multirow{3}{*}{\makecell[c]{Source \\ without \\ Patch}} & \multicolumn{2}{c|}{Training Methods ($\%$)} & \multicolumn{5}{c|}{Watermark Removal Approaches on CUPI ($\%$)} & \multicolumn{3}{c}{Average Drop$\uparrow$}\\
    \cline{2-11}
    & \makecell[c]{SL} & CUPI & FTAL~\cite{F/RTAL} & RTAL~\cite{F/RTAL} & EWC~\cite{EWC/AU} & AU~\cite{EWC/AU} & Overwriting & \multirow{2}{*}{CUPI} &\multirow{2}{*}{CUTI~\cite{CVPR}} & \multirow{2}{*}{NTL~\cite{NTL}}\\
    & \multicolumn{2}{c|}{[Test w/o Watermark ($\%$)]} & \multicolumn{5}{c|}{[Test w/o Watermark ($\%$)]}\\
    \midrule
    Digit~\cite{MT,US,SN,MM}                         & 96.7 $/$ 96.9 & \ 8.9 $/$ 96.0 & \ 8.6 $/$ \ 98.0 &  11.3 $/$ 97.1 &  10.2 $/$  100.0 & \ 8.7 $/$  100.0 & \ 9.8 $/$ 97.5 & \bf 88.8 & 87.5 & 82.6 \\
    CIFAR10 \& STL10~\cite{refCIFAR10,refSTL10}      & 85.6 $/$ 84.2 &  10.6 $/$ 86.6 &  11.1 $/$ \ 98.2 &  16.0 $/$ 97.7 &  14.2 $/$  100.0 &  11.6 $/$ \ 97.2 &  14.1 $/$ 92.4 & \bf 83.7 & 77.8 & 74.3  \\
    VisDA-2017~\cite{VisDA}                          & 93.6 $/$ 92.2 & \ 8.6 $/$ 95.2 &  19.9 $/$  100.0 &  22.2 $/$ 98.6 &  14.9 $/$  100.0 & \ 9.0 $/$  100.0 &  21.5 $/$ 98.8 & \bf 82.0 & 78.0 & 76.8  \\ 
    Office-Home-65~\cite{Home}                       & 75.5 $/$ 83.4 & \ 0.8 $/$ 83.1 & \ 1.3 $/$ \ 99.5 & \ 1.8 $/$ 99.5 & \ 1.7 $/$ \ 99.8 & \ 2.1 $/$ \ 99.6 & \ 3.4 $/$ 99.7 & \bf 97.6 & 94.5 & 94.1  \\ 
    DomainNet~\cite{DomainNet}                       & 39.0 $/$ 45.4 & \ 0.4 $/$ 44.7 & \ 3.2 $/$ \ 63.6 & \ 5.7 $/$ 59.3 & \ 0.8 $/$ \ 63.6 & \ 0.6 $/$ \ 62.9 & \ 3.2 $/$ 63.9 & \bf 60.0 & 52.2 & 47.3  \\       
    \bottomrule
  \end{tabular}}
  \label{ov_main}
\end{table*}

We conducted experiments on digital datasets, as depicted in Table~\ref{ts_digit_detail}. The vertical axis represents the authorized domain, while the horizontal axis represents the unauthorized domain. We explored 16 transfer tasks by selecting various combinations of authorized and unauthorized domains from the four domains of the digital datasets. In each task, the data on the left of '\(\Rightarrow\)' presents the test accuracy of supervised learning (SL) on the unauthorized domain, while the data on right of '\(\Rightarrow\)' presents the test accuracy of CUPI-Domain on the unauthorized domain. 'Authorized/Unauthorized Drop' indicates the drop (relative drop) of CUPI-Domain relative to SL in authorized/unauthorized domains. Table~\ref{ts_digit_compare} displays the 'Authorized/Unauthorized Drop' of CUPI-Domain, CUTI-Domain~\cite{CVPR}, and NTL~\cite{NTL}. From the results, we observe that the average drop of the CUPI-Domain on the unauthorized domain is 56.83 (86.89\%), while the drop on the authorized domain is only 0.02 (0.02\%).
Comparatively, CUTI-Domain~\cite{CVPR} and NTL~\cite{NTL} exhibit lower average performance degradations on the unauthorized domain but higher on the authorized domain. 
We further conduct statistical significance tests achieved by different methods, denoted with $^{\ast}$(CUPI-Domain vs. CUTI-Domain~\cite{CVPR}) and $^{\diamond}$(CUPI-Domain vs. NTL~\cite{NTL}). Our proposed CUPI-Domain shows a statistical difference (p $<$ 0.05) over CUTI-Domain~\cite{CVPR} and NTL~\cite{NTL}.
These results indicate that CUPI-Domain effectively reduces the sample recognition ability of the model for the unauthorized domain while having minimal impact on the authorized domain. 

Fig.~\ref{ts_csv} illustrates the performance on three different transfer tasks: CIFAR10 \(\rightarrow\) STL10, STL10 \(\rightarrow\) CIFAR10 and VisDA-T \(\rightarrow\) VisDA-V. Each subgraph displays the accuracy of the corresponding method in the authorized domain, the accuracy in the unauthorized domain, and their drop (relative drop).  SL achieves the highest classification accuracy on the unauthorized domain due to its wide generalization region. In contrast, the other three methods all reduced accuracy in the unauthorized domain, indicating the success of blocking the transfer pathway. Comparatively, CUPI-Domain shows higher degradation than others, demonstrating its superior ability to compress the model's generalization region.

To showcase the continued effectiveness of our method under increased numbers of categories and samples, we conducted experiments on Office-Home-65~\cite{Home} and DomainNet~\cite{DomainNet}, as shown in Supplementary Table~1,~2,~3 and~4. As the data complexity increases, we observed a decrease in the transfer performance of SL on the unauthorized domain (left of the \(\Rightarrow\)) compared to the results presented in Table~\ref{ts_digit_detail}. This decline poses a challenge to further reduce the accuracy of the unauthorized domain. However, our proposed CUPI-Domain demonstrates consistent effectiveness, with relative degradation of 83.41\% and 76.14\% on Office-Home-65~\cite{Home} and DomainNet~\cite{DomainNet}, respectively, exhibiting significantly better performance compared to CUTI-Domain~\cite{CVPR}. Despite the fact that NTL~\cite{NTL} exhibits a relatively higher degradation on DomainNet~\cite{DomainNet}, its performance experiences a significant drop of 16.06 (35.06\%) in the authorized domain, which is deemed unacceptable. Besides, our CUPI-Domain achieve statistical difference (p $<$ 0.05) over CUTI-Domain~\cite{CVPR} and NTL~\cite{NTL}.
This phenomenon can be attributed to the fact that the calculation of its maximum mean difference is class-independent, thereby reducing the model's feature recognition capability in the authorized domain to some extent, especially when the complexity of the dataset increases.

\begin{table*}[!t]
\renewcommand\arraystretch{1.4}
  \centering
    \caption{'Authorized/Unauthorized Drop' of target-free CUPI-Domain, CUTI-Domain~\cite{CVPR} and NTL~\cite{NTL} on digit datasets. 
    %The result of CUTI-Domain~\cite{CVPR} and NTL~\cite{NTL} is obtained by reproducing its source code for a fair comparison. The best performance is indicated by the numbers in bold. Statistical significance (p-value $<$ 0.05~\cite{P1,p2}) is denoted with: $^{\ast}$(CUPI-Domain vs. CUTI-Domain~\cite{CVPR}) and $^{\diamond}$(CUPI-Domain vs. NTL~\cite{NTL}).
    }
  \resizebox{0.9\textwidth}{!}{
  \begin{tabular}{c|ccc|ccc}
     \toprule
     \multirow{2}{*}{Domain / Drop} & \multicolumn{3}{c|}{Authorized Drop$\downarrow$} & \multicolumn{3}{c}{Unauthorized Drop$\uparrow$} \\
     \cline{2-7} 
     & CUPI-Domain & CUTI-Domain~\cite{CVPR} & NTL~\cite{NTL} & CUPI-Domain & CUTI-Domain~\cite{CVPR} & NTL~\cite{NTL} \\
    \midrule   
    MNIST~\cite{MT}   & \bf 0.03 (0.03\%) & 0.40 (0.40\%) & 0.70 (0.71\%) & \bf 60.43 (87.39\%) & 59.17 (85.43\%) & 57.30 (84.06\%) \\ 
    USPS~\cite{US}    & \bf 0.33 (0.33\%) & 0.60 (0.60\%) & 0.60 (0.60\%) & \bf 45.29 (81.80\%) & 44.97 (80.96\%) & 42.90 (74.71\%) \\
    SVHN~\cite{SN}    & \bf 1.10 (1.20\%) & 2.50 (2.74\%) & 3.20 (3.51\%) & \bf 44.87 (76.53\%) & 44.00 (74.19\%) & 41.53 (64.09\%) \\
    MNIST-M~\cite{MM} & \bf 0.19 (0.20\%) & 0.30 (0.31\%) & 2.00 (2.10\%) & \bf 69.82 (89.36\%) & 65.73 (83.96\%) & 63.03 (77.62\%) \\   
    \midrule
    Mean    & \bf 0.41 (0.44\%)$^{\ast \diamond}$ & 0.95 (1.02\%) & 1.63 (1.73\%) & \bf 55.10 (83.77\%)$^{\ast \diamond}$ & 53.47 (81.13\%) & 51.19 (75.12\%) \\
    \bottomrule
  \end{tabular}}
  \label{tf_dight_compare}
\end{table*}

\begin{figure*}[!t]
  \centering
   \includegraphics[width=0.9\linewidth,trim=5 185 5 185,clip]{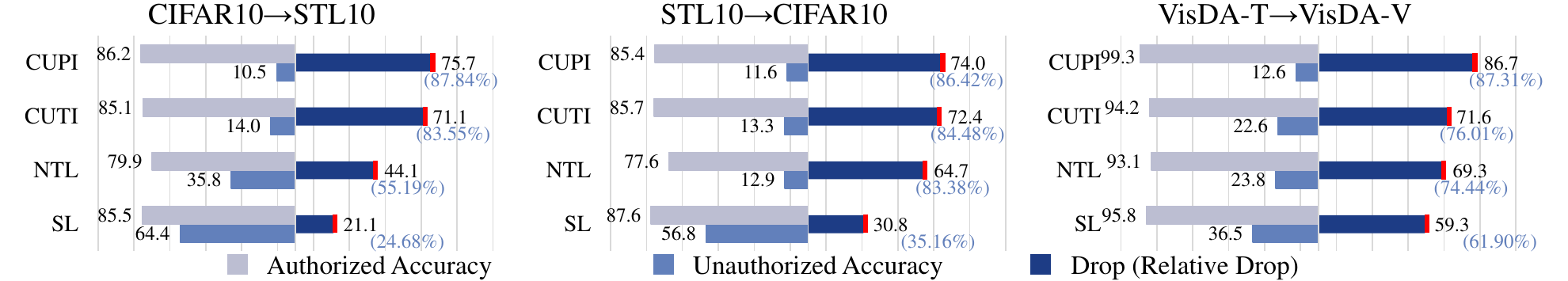}
   \caption{The accuracy ($\%$) of SL, target-free NTL~\cite{NTL}, CUTI-Domain~\cite{CVPR} and CUPI-Domain on CIFAR10\&STL10, and VisDA-2017. The title in each subgraph implies the authorized domain (left of '\(\rightarrow\)') and the unauthorized domain (right of '\(\rightarrow\)'). The bars with different colors represent the accuracy of each method in the authorized domain, the unauthorized domain, and the drop (relative drop) of the model performance, respectively.}
   \label{tf_csv}
\end{figure*}

\subsection{Result of Ownership Verification}
In this section, we focus on conducting ownership verification of the model by intentionally triggering classification errors. To this end, we apply a regular backdoor-based model watermark patch to the authorized domain dataset, followed by NTL~\cite{NTL}. 
This processed authorized domain is then treated as the new unauthorized domain. The accuracy of SL and CUPI-Domain on the authorized domain with/without the watermark patch is presented in Table~\ref{ov_main} and Supplemeatary Table~5.

Comparing the second and third columns of the table, we can observe that SL shows minimal disparity in accuracy with and with the watermark patch, indicating a lack of sensitivity to the watermark patch. Conversely, CUPI-Domain exhibits a significant reduction in accuracy on the unauthorized domain with watermark embedding. This performance difference can be effectively utilized for model ownership verification.

Additionally, we evaluate the robustness of CUPI-Domain against five state-of-the-art model watermark removal methods, including FTAL~\cite{F/RTAL}, RTAL~\cite{F/RTAL}, EWC~\cite{EWC/AU}, AU~\cite{EWC/AU}, and watermark overwriting followed by NTL~\cite{NTL}.
The last three columns depict the drop in accuracy between data with and without the watermark. Notably, CUPI-Domain, CUTI~\cite{CVPR}, and NTL~\cite{NTL} demonstrate effective resistance against watermark removal methods, with CUPI-Domain outperforming CUTI~\cite{CVPR} and NTL~\cite{NTL} by approximately 4.7\% and 8.2\% in terms of performance respectively, and also demonstrating statistical difference (p $<$ 0.05).

\subsection{Result of Target-free CUPI-Domain}
In Section 3.3.2, we proposed the target-free CUPI-Domain to address the scenario where the unauthorized domain is unknown by leveraging synthesized samples as a substitute for the unauthorized domain training set.
To evaluate the performance of our proposed target-free CUPI-Domain on digit datasets, we conducted 4 transfer tasks. For each task, we selected one authorized domain from the digit datasets and used the remaining unknown domains as the unauthorized domain testing sets. We report the 'Authorized/Unauthorized Drop' metric and statistical difference (p $<$ 0.05), as presented in Table~\ref{tf_dight_compare} (with more details in Supplementary Table~6). 
%The result of CUTI-Domain~\cite{CVPR} and NTL~\cite{NTL} is obtained by reproducing its source code for a fair comparison.

Analyzing the results, we observe that CUPI-Domain exhibits the highest average drop on the test sets, with a degradation of 55.10 (83.77\%). Additionally, CUPI-Domain demonstrates a lower drop in the unauthorized domain in comparison to CUTI-Domain~\cite{CVPR} and NTL~\cite{NTL}, further highlighting its enhanced IP protection capability.

Fig.~\ref{tf_csv} visually presents the outcomes for CIFAR10 \(\rightarrow\) STL10, STL10 \(\rightarrow\) CIFAR10 and VisDA-T \(\rightarrow\) VisDA-V. CUPI-Domain, CUTI~\cite{CVPR}, and NTL~\cite{NTL} exhibit higher degradation than SL consistent with the findings discussed earlier. Remarkably, CUPI-Domain demonstrates the highest degradation, indicating its superior ability to compress the model's generalization region. 
More experimental results on Office-Home-65~\cite{Home} and DomainNet~\cite{DomainNet} are detailed in Supplementary Table~7,~8,~9, and~10.

\subsection{Result of Applicability Authorization}

\begin{table*}
  \renewcommand\arraystretch{1.4}
  \centering
  \caption{The average 'Authorized Domain', 'Other Domains' and 'Drop' in applicability authorization scheme 1/2/3 of CUPI-Domain, CUTI-Domain~\cite{CVPR}, and NTL~\cite{NTL} on the digital datasets, CIFAR10\&STL10 and VisDA-2017. 
  %The result of CUTI-Domain~\cite{CVPR} and NTL~\cite{NTL} is obtained by reproducing its source code for a fair comparison. The best performance is indicated by the numbers in bold. Statistical significance (p-value $<$ 0.05~\cite{P1,p2}) is denoted with: $^{\ast}$(CUPI-Domain vs. CUTI-Domain~\cite{CVPR}) and $^{\diamond}$(CUPI-Domain vs. NTL~\cite{NTL}).
  }
  \resizebox{0.9\textwidth}{!}{
  
\begin{tabular}{c|ccc|ccc|ccc}
\toprule
\multirow{2}{*}{Scheme} & \multicolumn{3}{c|}{Authorized Doamin $\uparrow$} & \multicolumn{3}{c|}{Other Domains$\downarrow$} & \multicolumn{3}{c}{Drop$\uparrow$} \\
\cline{2-10}
& CUPI & CUTI~\cite{CVPR} & NTL~\cite{NTL} & CUPI & CUTI~\cite{CVPR} & NTL~\cite{NTL} & CUPI & CUTI~\cite{CVPR} & NTL~\cite{NTL} \\
\midrule
Scheme-1 & \bf \ 99.84$^{\ast \diamond}$  &   \ 99.37 &      99.03 & \bf  11.27$^{\ast \diamond}$ & 15.62 & 18.00 & \bf 88.57 (88.71\%)$^{\ast \diamond}$ & 83.75 (84.27\%) & 81.03 (81.81\%) \\  
Scheme-2 & \bf \ 99.75$^{\ast \diamond}$ &    \ 99.47 &      99.34 & \bf  14.25$^{\ast \diamond}$ & 16.63 & 18.60 & \bf 85.50 (85.71\%)$^{\ast \diamond}$ & 85.85 (83.27\%) & 80.75 (81.28\%) \\  
Scheme-3 & \bf \ 99.08$^{\ast \diamond}$ &    \ 98.77 &      98.64 & \bf  12.72$^{\ast \diamond}$ & 14.78 & 16.90 & \bf 86.36 (87.16\%)$^{\ast \diamond}$ & 84.00 (85.05\%) & 81.74 (82.88\%) \\  

\bottomrule
\end{tabular}
}
  \vskip -10pt
  \label{aa_all}
\end{table*}

To assess the applicability authorization, we conduct experiments by constraining its generalization capabilities solely to the authorized domain. We adopted three evaluation schemes: 
1) We introduce a specified watermark patch to the authorized domain (same as in Section 4.4), creating a new authorized domain training set. Subsequently, we combine the authorized domain, synthesized domain, and synthesized domain with the specified watermark patch to construct the unauthorized domain training set. 
2) We use clean authorized data without a watermark patch as an authorized domain training set and then combine the authorized domain with a watermark patch, synthesized domain, and synthesized domains with a watermark patch to construct the unauthorized domain training set.
3) We consider clean authorized data without a watermark patch as an authorized domain training set and the synthesized domain as the unauthorized domain training set.
During the testing phase, we evaluated the model's performance on several unknown domains, both with and without specified watermark patches.

The experimental results are reported in Table~\ref{aa_all}, with additional details available in Supplementary Table~11,~12,~13,~14,~15 and~16. The results of scheme 1 indicate that the proposed CUPI-Domain performs well on the authorized domain with the specific watermark patch but exhibits low accuracy on other unknown domains. In scheme 2, the proposed CUPI-Domain performs well only on the authorized domain without the watermark patch. 
When the watermark is considered insufficient to change the domain in scheme 3, the CUPI-Domain performs effectively on authorized domains, regardless of the presence of watermarks.
This aligns with our expectation that the model's generalization ability is successfully constrained within the authorized domain, regardless of the presence or absence of the watermark patch.

Furthermore, in scheme 1/2/3, our proposed CUPI-Domain achieves an average drop of 88.57 (88.71\%) / 85.50 (85.71\%) / 86.36 (87.16\%), surpassing CUTI's drop (relative drop) of 83.75 (84.27\%) / 85.85 (83.27\%) / 84.00 (85.05\%) and NTL's drop (relative drop) of 81.03 (81.81\%) / 80.75 (81.28\%) / 81.74 (82.88\%) with a statistical difference (p $<$ 0.05). This improvement is attributed to the construction of the CUPI-Domain, where we leverage infinite samples that closely resemble the authorized domain to create a more compact generalization boundary. Consequently, CUPI-Domain exhibits a stronger ability to protect IP compared to NTL~\cite{NTL}, which relies solely on limited features from the source and target domains for distance maximization.

\begin{table*}[!t]
\renewcommand\arraystretch{1.4}
  \centering
    \caption{Accuracy ($\%$) of CUPI-Domain on five random tasks with various components combinations. The vertical axis axis represents the combinations of different components, while the horizontal axis represents the loss function, with '$\checkmark$' indicating the correspondence. 
    %The best performance is indicated by the numbers in bold.
    }
  \resizebox{0.9\textwidth}{!}{
  \begin{tabular}{l|ccccc|ccccc}
    \toprule
    \multirow{3}{*}{Architecture} & \multirow{3}{*}{\({\cal L}_s\)} & \multirow{3}{*}{\({\cal L}_i\)} & \multirow{3}{*}{\({\cal L}_t\)} & \multirow{3}{*}{\({{\cal L}_{Stl}}\)} & \multirow{3}{*}{\({{\cal L}_{Dis}}\)} & \multirow{3}{*}{\begin{tabular}[c]{@{}c@{}} MNIST \\ \(\rightarrow\) \\ SVHN \end{tabular}} & \multirow{3}{*}{\begin{tabular}[c]{@{}c@{}} CIFAR10 \\ \(\rightarrow\) \\ STL10 \end{tabular}} & \multirow{3}{*}{\begin{tabular}[c]{@{}c@{}} VisDA-T \\ \(\rightarrow\) \\ VisDA-V \end{tabular}} & \multirow{3}{*}{\begin{tabular}[c]{@{}c@{}} Artistic \\ \(\rightarrow\) \\ Product \end{tabular}} & \multirow{3}{*}{\begin{tabular}[c]{@{}c@{}} quickdraw \\ \(\rightarrow\) \\ infograph \end{tabular}} \\  
    &  & \multicolumn{1}{c}{} & \multicolumn{1}{c}{} & \multicolumn{1}{c}{} & \multicolumn{1}{c|}{} & \multicolumn{1}{c}{} & \multicolumn{1}{c}{} & \multicolumn{1}{c}{} & \multicolumn{1}{c}{} \\ 
    &  & \multicolumn{1}{c}{} & \multicolumn{1}{c}{} & \multicolumn{1}{c}{} & \multicolumn{1}{c|}{} & \multicolumn{1}{c}{} & \multicolumn{1}{c}{} & \multicolumn{1}{c}{} & \multicolumn{1}{c}{} \\ 
    \midrule
    SL                                                             & \checkmark & & & & & 38.2 & 64.4 & 36.5 & 64.9 & 3.5 \\ 
    SL+CUPI-Domain generators (without \({\cal L}_t\))             & \checkmark & \checkmark & & & & 11.4 & 10.1 & 10.7 & 10.9 & 2.2 \\ 
    SL+CUPI-Domain generators (without \({\cal L}_i\))             & \checkmark & & \checkmark & & & 9.4 & 11.4 & 11.9 & 10.4 & 3.0  \\ 
    SL+CUPI-Domain generators                                      & \checkmark & \checkmark & \checkmark & & & 9.4 & 11.2 & 10.1 & 9.9 & 2.8  \\ 
    SL+CUPI-Domain generators+DIMB (without \({{\cal L}_{Stl}}\))  & \checkmark & \checkmark & \checkmark & \checkmark & & 6.3 & 11.0 & 8.2 & 9.4 & 1.8  \\ 
    SL+CUPI-Domain generators+DIMB (without \({{\cal L}_{Dis}}\))  & \checkmark & \checkmark & \checkmark & & \checkmark & 6.3 & 9.9 & 9.6 & 9.4 & 1.6  \\ 
    SL+CUPI-Domain generators+DIMB (CUPI-Domain)                   & \checkmark & \checkmark & \checkmark & \checkmark & \checkmark & \bf 6.1 & \bf 9.5 & \bf 7.5 & \bf 8.1 & \bf 1.4  \\ 
    \bottomrule
  \end{tabular}}
  \label{loss}
\end{table*}

\begin{figure*}[!t]
  \centering
   \includegraphics[width=0.9\linewidth,trim=40 165 40 150,clip]{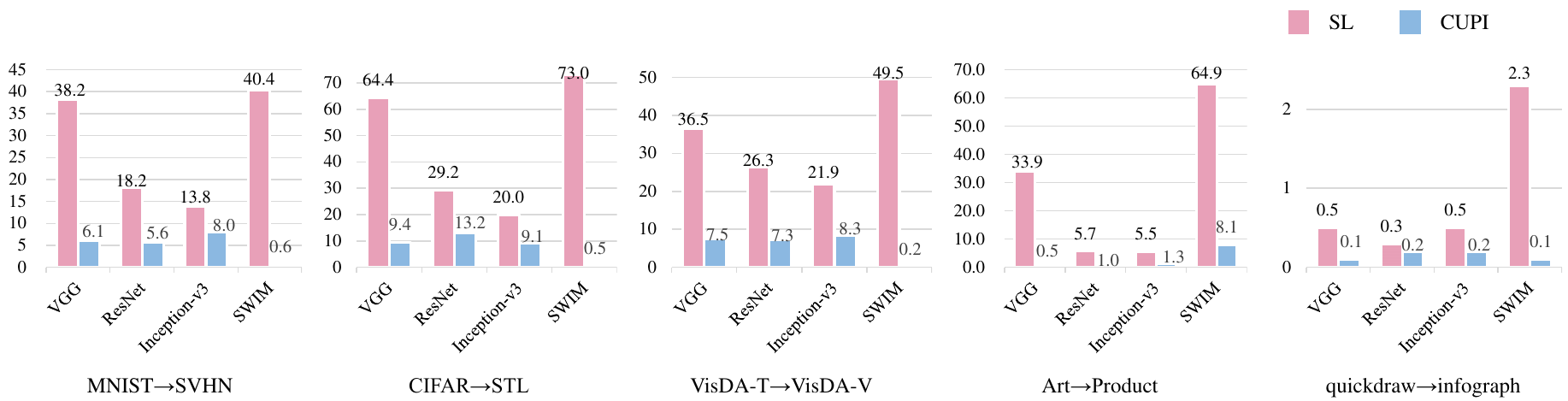}
   \caption{The accuracy ($\%$) of SL and target-specified CUPI-Domain combined with four different backbones (VGG\cite{VGG}, ResNet~\cite{ResNet}, Inception-v3~\cite{Inception}, and SWIM~\cite{SWIN} on five random tasks.}
   \label{backbone}
\end{figure*}

\subsection{Ablation Study of Components}

To gain further insights into the effectiveness of each component in our proposed CUPI-Domain, we conduct an ablation study on five random tasks, as detailed in Table~\ref{loss}.
The vertical axis represents experimental settings with different component combinations (indicated by '+'), and the checkmark '$\checkmark$' indicates the corresponding loss function on the horizontal axis.
For each task, we randomly selected two domains from each dataset, designating one as the authorized domain (left of '\(\rightarrow\)') and the other as the unauthorized domain (right of '\(\rightarrow\)'), as shown on the horizontal axis.

The complexity of the dataset varies, leading to differences in the difficulty of feature extraction and the construction of the CUPI-Domain. Consequently, the accuracy of CUPI-Domain with different partial combinations on the unauthorized domain also varies.
However, three key observations remain consistent across all tasks: 
firstly, the accuracy of different variants on the unauthorized domain is significantly lower than that of SL, indicating the positive impact of each component; secondly, the variants with DIMB perform better than those without it, indicating that DIMB can obtain stable domain class features and domain class-wise style features, further amplifying the difference between authorized domains and unauthorized domains, thereby reducing the identification of unauthorized domains. thirdly, the complete combinations consistently achieve the lowest accuracy on the unauthorized domain, highlighting the effectiveness of our proposed loss function in protecting model IP.

\subsection{Ablation Study of Backbone}

In this section, we investigate the IP protection capability of the proposed CUPI-Domain in combination with different backbone architectures, as illustrated in Fig.~\ref{backbone}. 
We selected four distinct backbone networks, including VGG\cite{VGG}, ResNet~\cite{ResNet}, Inception-v3~\cite{Inception}, and SWIM~\cite{SWIN}.
We perform a total of five transfer tasks based on the number of available datasets. For each transfer task, we randomly selected two domains from each dataset, designating one as the authorized domain (left of '\(\rightarrow\)') and the other as the unauthorized domain (right of '\(\rightarrow\)'). In each subfigure, the pink bars indicate the accuracy of SL on the unauthorized domain, while the blue bars depict the accuracy of CUPI-Domain on the unauthorized domain.

Based on the experimental results, we can observe a consistent pattern: irrespective of the dataset, CUPI-Domain consistently outperforms SL in reducing the recognition ability on the unauthorized domain when integrated with various backbones.
Furthermore, when combined with SWIM\cite{SWIN}, CUPI-Domain exhibits higher accuracy on the authorized domain and lower accuracy on the unauthorized domain. This can be attributed to the stronger feature extraction ability of SWIM\cite{SWIN} on complex datasets compared to the other backbones. As a result, when integrated with SWIM\cite{SWIN}, CUPI-Domain becomes more similar to the authorized domain, leading to a more compact representation of the model's performance within the authorized domain. This signifies a stronger IP protection ability with a stronger feature extractor.

\section{Discussion}
\subsection{Protection \& Generalization}
In the target-specified scenario, CUPI-Domain effectively diminishes the identification ability of unauthorized domains by constraining the generalization boundary of the authorized domain while preserving its universality. However, in target-free scenarios, where the distribution of unauthorized domains varies, the IP protection mechanism of the model naturally results in a certain degree of loss of universality.

The trade-off between protection and generalization is a critical consideration. While our framework prioritizes model IP protection, it is equally imperative to ensure that the protected model remains effective across diverse deployment scenarios. We acknowledge the need for further investigation and research to address this trade-off, such as exploring methods to enhance the generalization ability of protected models through additional training or data augmentation strategies. This area is one that we plan to delve into in future work to ensure that our framework remains practical and effective in real-world applications.

\subsection{Attacks from Input Data}
The design of our proposed CUPI-Domain is grounded in the deep features of the domain, extracted by the feature extractor of the backbone. By emphasizing the private features of authorized domains, CUPI-Domain can establish more restrictive generalization boundaries for trained models, implicitly preventing unauthorized transfer to unauthorized domains with irrelevant private-style features. In this section, we further explore whether CUPI-Domain can resist input-based IP attacks.

Recently, several image translation methods based on CycleGAN~\cite{CycleGAN} have been proposed to translate images between two different domains, aiming to reduce the domain gap~\cite{CycleGANattack}. To verify the resistance of CUPI-Domain to such methods, we applied CycleGAN~\cite{CycleGAN} to generate transformed images for each task pair (i.e., any two different domains with the same task) and tested the corresponding trained CUPI-Domain, CUTI-Domain~\cite{CVPR} and NTL~\cite{NTL} on these images to assess their IP protection capabilities. The experimental results of the trained target-specified model and target-free model are presented in Table~\ref{attacks} (with more details in Supplementary Table~17).

Table~\ref{attacks} showcases the average accuracy of the trained target-specified/free models on translated images. The mean accuracy of different methods ranges from 12\% to 25\%, demonstrating poor recognition ability. These results suggest a certain level of resistance to input-based attacks of these IP protection methods, with CUPI-Domain demonstrating the best performance with a statistical difference (p $<$ 0.05). This underscores the effectiveness of CUPI-Domain in preserving the IP of the model against potential adversarial attacks.

\begin{table}[!t]
\renewcommand\arraystretch{1.4}
  \centering
    \caption{The average accuracy ($\%$) of target-specified/free CUPI-Domain, CUTI-Domain~\cite{CVPR} and NTL~\cite{NTL} on translated input data~\cite{CycleGAN}, the unauthorized domain after model stealing attack by DeepSteal~\cite{Steal1} and Knockoff-Nets~\cite{Steal2}. 
    % The best performance is indicated by the numbers in bold. Statistical significance (p-value $<$ 0.05~\cite{P1,p2}) is denoted with: $^{\ast}$(CUPI-Domain vs. CUTI-Domain~\cite{CVPR}) and $^{\diamond}$(CUPI-Domain vs. NTL~\cite{NTL}).
    }
  \resizebox{0.48\textwidth}{!}{
  \begin{tabular}{c|ccc|ccc}
    \toprule
    \multirow{2}{*}{Domain} & \multicolumn{3}{c|}{Target-Specified} & \multicolumn{3}{c}{Target-Free} \\
    \cline{2-7}
    & CUPI & CUTI~\cite{CVPR} & NTL~\cite{NTL} & CUPI & CUTI~\cite{CVPR} & NTL~\cite{NTL} \\
    \midrule   
    CycleGAN~\cite{CycleGAN} & \bf 12.39$^{\ast}$  & 15.45  & 15.97 & \bf 21.06$^{\ast}$  & 22.41  & 24.05 \\
    DeepSteal~\cite{Steal1} & \bf 8.71$^{\diamond}$  & 9.45  & 13.53  & \bf 10.19$^{\ast \diamond}$  & 11.31  & 17.51 \\
    Knockoff~\cite{Steal2} & \bf 10.75$^{\diamond}$  & 16.80  & 17.19  & \bf 9.57$^{\ast \diamond}$  & 19.92  & 17.08  \\
    \bottomrule
  \end{tabular}}
  \label{attacks}
\end{table}

\subsection{Attacks from Model Stealing}
We further verify the ability of the proposed CUPI-Domain to resist some of the latest model stealing attack methods (i.e., DeepSteal~\cite{Steal1} and Knockoff-Nets~\cite{Steal2}). Table~\ref{attacks}, Supplementary Table~18 and~19 show the accuracy of the substitute model according to CUPI-Domain, CUTI-Domain~\cite{CVPR}, and NTL~\cite{NTL} on the unauthorized domain in target-specified/free scenarios. The average accuracy of the unauthorized domain remains low, indicating that these methods can effectively thwart model-stealing attacks to some extent. Among them, CUPI-Domain has the lowest accuracy with a statistical difference (p $<$ 0.05), demonstrating the strongest IP protection ability.

\subsection{Exploration on More Data Types}
To further verify the applicability and limitations of the proposed CUPI-Domain, we deployed it on text classification tasks (5 categories, random 4 domains from 35)~\cite{Amazon} and audio classification tasks (10 categories, 3 domains)~\cite{Audio}. The experimental results are presented in Table~\ref{types}, Supplementary Table~20 and Table~21. As shown in the Table, CUPI-Domain, CUTI-Domain~\cite{CVPR}, and NTL~\cite{NTL} have successfully reduced the performance of unauthorized access while retaining the performance of authorized domains, demonstrating certain IP protection capabilities, with CUPI-Domain shows the strongest protection capability with a statistical difference (p $<$ 0.05).

However, it should be emphasized that the design and development of CUPI-Domain are based on image features, and its performance on text and audio tasks is weaker than on images. Further exploration and improvement are needed to enhance CUPI-Domain's adaptability to text and audio IP protection tasks. This is an area that we plan to explore in future work to ensure that our framework remains adaptable and effective in more real-world applications.

\begin{table*}[!t]
\renewcommand\arraystretch{1.4}
  \centering
    \caption{Results of target-specified CUPI-Domain, CUTI-Domain~\cite{CVPR} and NTL~\cite{NTL} on Amazon Reviews Dataset~\cite{Amazon} and DCASE Dataset~\cite{Audio}.}
  \resizebox{0.9\textwidth}{!}{
  \begin{tabular}{c|ccc|ccc}
     \toprule
     \multirow{2}{*}{Domain / Drop} & \multicolumn{3}{c|}{Authorized Drop$\downarrow$} & \multicolumn{3}{c}{Unauthorized Drop$\uparrow$} \\
     \cline{2-7} 
     & CUPI-Domain & CUTI-Domain~\cite{CVPR} & NTL~\cite{NTL} & CUPI-Domain & CUTI-Domain~\cite{CVPR} & NTL~\cite{NTL} \\
    \midrule   
    Amazon~\cite{Amazon}                & \bf 0.17 (0.26\%)$^{\ast \diamond}$ & 0.56 (0.90\%) & 0.53 (0.82\%) & \bf 42.50 (70.13\%)$^{\ast \diamond}$ & 40.26 (66.35\%) & 39.94 (65.76\%) \\
    DCASE~\cite{Audio}     & \bf 0.00 (0.00\%)$^{\diamond}$ & 0.17 (0.32\%) & 0.87 (1.62\%) & \bf 16.67 (58.17\%)$^{\ast \diamond}$ & 14.76 (48.63\%) & 14.06 (46.25\%) \\
    \bottomrule
  \end{tabular}}
  \label{types}
\end{table*}

\subsection{Exploration on Real-World Clinical Data}
To further validate the applicability of the proposed CUPI-Domain in real-world scenarios, we deployed it in a tuberculosis classification task with two categories and two domains (ChinaSet~\cite{refc} and Montgomery CXR~\cite{refg}). The experimental results are presented in Table~\ref{ct_all} (with more details in Supplementary Table~22, Supplementary Table~23), which demonstrates that CUPI-Domain exhibits robust IP protection capabilities in both target-specified and target-free scenarios, with a statistical difference (p $<$ 0.05).

\begin{table*}[!t]
\renewcommand\arraystretch{1.4}
  \centering
    \caption{Results of target-specified/free CUPI-Domain, CUTI-Domain~\cite{CVPR} and NTL~\cite{NTL} on the tuberculosis classification task. 
    %The best performance is indicated by the numbers in bold. Statistical significance (p-value $<$ 0.05~\cite{P1,p2}) is denoted with: $^{\ast}$(CUPI-Domain vs. CUTI-Domain~\cite{CVPR}) and $^{\diamond}$(CUPI-Domain vs. NTL~\cite{NTL}).
    }
  \resizebox{0.9\textwidth}{!}{
  \begin{tabular}{c|ccc|ccc}
    \toprule
    \multirow{2}{*}{Domain / Drop} & \multicolumn{3}{c|}{Authorized Drop$\downarrow$} & \multicolumn{3}{c}{Unauthorized Drop$\uparrow$} \\
    \cline{2-7}
    & CUPI & CUTI~\cite{CVPR} & NTL~\cite{NTL} & CUPI & CUTI~\cite{CVPR} & NTL~\cite{NTL} \\
    \midrule
    Target-specified & \bf 0.00 (0.00\%)$^{\ast \diamond}$ & 0.50 (0.55\%) & 0.50 (0.55\%) & 51.05 (75.88\%)$^{\ast \diamond}$ & 46.85 (69.39\%) & 46.85 (70.50\%) \\
    Target-free & \bf 0.00 (0.00\%)$^{\ast \diamond}$ & 1.05 (1.17\%) & 1.05 (1.17\%) & 45.30 (66.85\%)$^{\ast \diamond}$ & 42.15 (61.99\%) & 37.45 (54.86\%) \\
    \bottomrule
  \end{tabular}}
  \label{ct_all}
\end{table*}

	% 0.00 (0.00\%) & 0.00 (0.00\%) & 0.00 (0.00\%) & 62.50 (80.03\%) & 58.30 (74.65\%) & 54.10 (69.27\%)  \\ 
 %        0.00 (0.00\%) & 1.00 (1.10\%) & 1.00 (1.10\%) & 39.60 (71.74\%) & 35.40 (64.13\%) & 39.60 (71.74\%)  \\ 
 %        0.00 (0.00\%) & 0.50 (0.55\%) & 0.50 (0.55\%) & 51.05 (75.88\%) & 46.85 (69.39\%) & 46.85 (70.50\%)  \\ 
 %        0.00 (0.00\%) & 1.10 (1.23\%) & 2.10 (2.34\%) & 57.30 (73.37\%) & 54.10 (69.27\%) & 48.90 (62.61\%)  \\ 
 %        0.00 (0.00\%) & 1.00 (1.10\%) & 0.00 (0.00\%) & 33.30 (60.33\%) & 30.20 (54.71\%) & 26.00 (47.10\%)  \\ 
 %        0.00 (0.00\%) & 1.05 (1.17\%) & 1.05 (1.17\%) & 45.30 (66.85\%) & 42.15 (61.99\%) & 37.45 (54.86\%)  \\ 

\section{Conclusion}
Protecting model intellectual property (IP) in the field of artificial intelligence is a significant and crucial challenge, and its significance cannot be overstated.
To tackle this issue, we propose a novel approach called CUPI-Domain, which acts as a protective barrier to confine the model's performance within authorized domains.
To achieve this, the pyramid CUPI-Domain generator is designed to construct an isolation domain that closely resembles the features of the authorized domain, thereby hindering the model's ability to generalize to unauthorized domains and leading to recognition failure when encountering new private-style features.
Additionally, we design external DIMB for storing and updating labeled pyramid features with corresponding loss functions.
Furthermore, we offer two solutions to deploy the CUPI-Domain: target-specified and target-free, depending on whether information about the unauthorized domain is available. Through extensive experiments conducted on several popular cross-domain datasets, we demonstrate the effectiveness of our lightweight and easily deployable CUPI-Domain. We believe that our work will contribute to the advancement of research on model IP protection and security, highlighting its crucial importance in real-world applications. 

\section*{Acknowledgments}
This work was supported by the National Natural Science Foundation of China (Nos. 62136004, 62276130), the Key Research and Development Plan of Jiangsu Province (No. BE2022842), and Huazhu Fu's A*STAR Central Research Fund and Career Development Fund (C222812010).

\ifCLASSOPTIONcaptionsoff
  \newpage
\fi

\bibliographystyle{IEEEtran}  % Specify the desired BibTeX style
{\small 
\bibliography{ref}
}

\clearpage
\setcounter{table}{0} 
\setcounter{figure}{0} 
\section*{Supplementary Material}
% \newcolumntype{L}{>{\hspace{0.1in}\arraybackslash}p{0.4\textwidth}}
% \newcolumntype{C}{>{\centering\arraybackslash}p{0.4\textwidth}}

\title{Supplementary Material}
\maketitle

\begin{minipage}[!t]{0.9\textwidth}
% \begin{minipage}[t]{\textwidth}
\begin{algorithm}[H]
\caption{Target-Specified CUPI-Domain.}\label{alg:alg1}
\begin{algorithmic}[1]
\REQUIRE{The authorized domain \(X_s\), CUPI-Domain \(X_i\), the unauthorized domain \(X_t\), number of feature extractor blocks \(L\), number of epoch \(E\), the backbone parameters $\theta$, and the CUPI-Domain generator parameters $\phi$.}
\STATE $ \text{Initialize CUPI-Domain with unauthorized domain.} $
\STATE $ \text{Initialize DIMB \(M^1, M^2,…,M^L\) and \(M^D\).} $
\STATE $ \textbf{For \(epoch=1\) to \(E\) do} $
\STATE $ \textbf{\qquad For \(l=1\) to \(L\) do} $
\STATE $ \text{\qquad \qquad Calculate the output of \(X_s\), \(X_i\) and \(X_t\) in the \(l\)-th feature block: \(f_s^l\), \(f_i^l\) and \(f_t^l\).} $
\STATE $ \text{\qquad \qquad Update \(M^l\).} $
\STATE $ \text{\qquad \qquad Update \(f_i^l\) according to Eq.~(2).} $
\STATE $ \textbf{\qquad End For} $
\STATE $ \text{\qquad Update \(M^D\).} $
\STATE $ \text{\qquad Update backbone parameters $\theta$ and CUPI-Domain generator parameters $\phi$ by Eq.~(3).} $
\STATE $ \textbf{End For} $
\STATE Return backbone parameters $\theta$ and and the CUPI-Domain generator parameters $\phi$.
\end{algorithmic}
\label{alg1}
\end{algorithm}
\end{minipage}

\begin{minipage}[!t]{0.9\textwidth}
\begin{algorithm}[H]
\caption{Target-Free CUPI-Domain.}\label{alg:alg1}
\begin{algorithmic}[1]
\REQUIRE{The authorized domain \(X_s\), number of feature extractor blocks \(L\), number of epoch \(E\), the backbone parameters $\theta$, and the CUPI-Domain generator parameters $\phi$.}
\STATE $ \text{Generate synthesized unauthorized domain by } $
$ \text{ Wang~\textit{et al.}~[14] and Huang~\textit{et al.}~[50].} $
\STATE $ \text{Initialize CUPI-Domain with unauthorized domain.} $
\STATE $ \text{Initialize  DIMB \(M^1, M^2,…,M^L\) and \(M^D\).} $
\STATE $ \textbf{For \(epoch=1\) to \(E\) do} $
\STATE $ \textbf{\qquad For \(l=1\) to \(L\) do} $
\STATE $ \text{\qquad \qquad Calculate the output of \(X_s\), \(X_i\) and \(X_t\) in the \(l\)-th feature block: \(f_s^l\), \(f_i^l\) and \(f_t^l\).} $
\STATE $ \text{\qquad \qquad Update \(M^l\).} $
\STATE $ \text{\qquad \qquad Update \(f_i^l\) according to Eq.~(2).} $
\STATE $ \textbf{\qquad End For} $
\STATE $ \text{\qquad Update \(M^D\).} $
\STATE $ \text{\qquad Update backbone parameters $\theta$ and CUPI-Domain generator parameters $\phi$ by Eq.~(3).} $ 
\STATE $ \textbf{End For} $
\STATE Return backbone parameters $\theta$ and and the CUPI-Domain generator parameters $\phi$.
\end{algorithmic}
\label{alg2}
\end{algorithm}
\end{minipage}

\begin{table*}[!t]
\renewcommand\arraystretch{1.6}
  \centering
    \caption{The accuracy ($\%$) of target-specified CUPI-Domain on the Office-Home-65~[28]. The vertical/horizontal axis denotes the authorized/unauthorized domain. In each task, the left of '\(\Rightarrow\)' shows the test accuracy of supervised learning (SL) on the unauthorized domain, while the right side presents the accuracy of CUPI-Domain. 'Authorized/Unauthorized Drop' indicate the drop (relative drop) of CUPI-Domain relative to SL on authorized/unauthorized domains.}
  \resizebox{1\textwidth}{!}{
  \begin{tabular}{c|cccc|cc}
    \toprule
    Authorized/Unauthorized & Artistic & Clipart & Product & RealWorld & Authorized Drop$\downarrow$ & Unauthorized Drop$\uparrow$ \\
    \midrule   
    Artistic  & 76.3 $\Rightarrow$  76.3 & 47.1 $\Rightarrow$ \ 2.1 & 64.9 $\Rightarrow$ \ 8.1 & 72.2 $\Rightarrow$  26.3 & 0.00 (0.00\%) & 49.25 (82.23\%) \\ 
    Clipart   & 57.8 $\Rightarrow$ \ 2.3 & 80.1 $\Rightarrow$  79.9 & 63.5 $\Rightarrow$ \ 2.6 & 68.8 $\Rightarrow$ \ 4.9 & 0.20 (0.25\%) & 60.05 (94.88\%) \\ 
    Product   & 56.6 $\Rightarrow$ \ 7.8 & 45.2 $\Rightarrow$ \ 2.3 & 92.7 $\Rightarrow$  92.4 & 72.7 $\Rightarrow$  22.1 & 0.30 (0.32\%) & 47.38 (83.52\%) \\ 
    RealWorld & 63.8 $\Rightarrow$  32.0 & 49.2 $\Rightarrow$ \ 4.4 & 75.5 $\Rightarrow$  16.4 & 84.4 $\Rightarrow$  84.4 & 0.00 (0.00\%) & 45.21 (73.02\%) \\         
    \bottomrule
  \end{tabular}}
  \label{ts_home_detail}
\end{table*}

\begin{table*}[!t]
\renewcommand\arraystretch{1.6}
  \centering
    \caption{'Authorized/Unauthorized Drop' of target-specified CUPI-Domain, CUTI-Domain~[27] and NTL~[14] on the Office-Home-65~[28]. The result of CUTI-Domain~[27] and NTL~[14] is obtained by reproducing its source code for a fair comparison.
    The best performance is indicated by the numbers in bold. Statistical significance (p-value $<$ 0.05~[60],~[61]) is denoted with: $^{\ast}$(CUPI-Domain vs. CUTI-Domain~[27]) and $^{\diamond}$(CUPI-Domain vs. NTL~[14]).}
  \resizebox{1\textwidth}{!}{
  \begin{tabular}{c|ccc|ccc}
    \toprule
    \multirow{2}{*}{Domain / Drop} & \multicolumn{3}{c|}{Authorized Drop$\downarrow$} & \multicolumn{3}{c}{Unauthorized Drop$\uparrow$} \\
    \cline{2-7} 
     & CUPI-Domain & CUTI-Domain~[27] & NTL~[14] & CUPI-Domain & CUTI-Domain~[27] & NTL~[14] \\
    \midrule   
    Artistic  & \bf 0.00 (0.00\%) &     0.30 (0.39\%) &     0.80 (1.05\%) & \bf 49.25 (82.23\%) & 47.16 (78.77\%) & 37.27 (65.98\%) \\ 
    Clipart   & \bf 0.20 (0.25\%) & \bf 0.20 (0.25\%) & \bf 0.20 (0.25\%) & \bf 60.05 (94.88\%) & 57.35 (90.66\%) & 54.31 (86.21\%) \\ 
    Product   & \bf 0.30 (0.32\%) &     0.50 (0.54\%) & \bf 0.30 (0.32\%) & \bf 47.38 (83.52\%) & 45.82 (80.69\%) & 45.01 (80.16\%) \\ 
    RealWorld & \bf 0.00 (0.00\%) &     0.30 (0.36\%) &     2.40 (2.84\%) & \bf 45.21 (73.02\%) & 42.95 (69.11\%) & 30.37 (53.31\%) \\         
    \midrule
    Mean      & \bf 0.13 (0.14\%)$^{\ast \diamond}$ &     0.33 (0.38\%) &     0.43 (1.12\%) & \bf 50.47 (83.41\%)$^{\ast \diamond}$ & 48.32 (79.79\%) & 41.74 (71.42\%) \\
    \bottomrule
  \end{tabular}}
  \label{ts_home_compare}
\end{table*}

\begin{table*}[!t]
\renewcommand\arraystretch{1.6}
  \centering
    \caption{The accuracy ($\%$) of target-specified CUPI-Domain on the DomainNet~[29]. The vertical/horizontal axis denotes the authorized/unauthorized domain. In each task, the left of '\(\Rightarrow\)' shows the test accuracy of SL on the unauthorized domain, while the right side presents the accuracy of CUPI-Domain. 'Authorized/Unauthorized Drop' indicate the drop (relative drop) of CUPI-Domain relative to SL on authorized/unauthorized domains.}
  \resizebox{1\textwidth}{!}{
  \begin{tabular}{c|cccccc|cc}
    \toprule
    Authorized/Unauthorized & clipart & infograph & painting & quickdraw & real & sketch & Authorized Drop$\downarrow$ & Unauthorized Drop$\uparrow$ \\
    \midrule   
    clipart   & 67.2 $\Rightarrow$  62.2 &  12.8 $\Rightarrow$ \ 3.3 &  29.7 $\Rightarrow$ \ 8.3 &  13.8 $\Rightarrow$ \ 0.0 & 51.2 $\Rightarrow$  20.4 &  31.6 $\Rightarrow$  15.5 & 4.99 ( \ 7.42\%) &  18.34 (71.44\%) \\
    infograph & 21.2 $\Rightarrow$ \ 7.1 &  23.0 $\Rightarrow$  22.2 &  21.1 $\Rightarrow$ \ 6.4 & \ 3.6 $\Rightarrow$ \ 0.0 & 35.1 $\Rightarrow$ \ 6.8 &  16.2 $\Rightarrow$ \ 2.8 & 0.80 ( \ 3.48\%) &  14.83 (79.79\%) \\
    painting  & 28.9 $\Rightarrow$  10.0 &  13.6 $\Rightarrow$ \ 4.9 &  50.0 $\Rightarrow$  49.9 & \ 3.7 $\Rightarrow$ \ 0.0 & 45.6 $\Rightarrow$  35.2 &  22.1 $\Rightarrow$ \ 5.3 & 0.14 ( \ 0.49\%) &  11.74 (65.61\%) \\
    quickdraw & 17.2 $\Rightarrow$ \ 0.1 & \ 2.3 $\Rightarrow$ \ 0.1 & \ 6.1 $\Rightarrow$ \ 0.0 &  51.0 $\Rightarrow$  44.1 & 12.6 $\Rightarrow$ \ 0.0 & \ 8.6 $\Rightarrow$ \ 0.3 & 6.95 (  13.62\%) & \ 9.23 (97.73\%) \\
    real      & 39.6 $\Rightarrow$  17.3 &  13.4 $\Rightarrow$ \ 6.3 &  40.3 $\Rightarrow$  20.1 & \ 6.4 $\Rightarrow$ \ 0.0 & 78.1 $\Rightarrow$  69.9 &  25.8 $\Rightarrow$ \ 6.8 & 8.15 (  10.44\%) &  15.01 (66.65\%) \\
    sketch    & 35.8 $\Rightarrow$  15.8 &  11.3 $\Rightarrow$ \ 2.9 &  25.7 $\Rightarrow$ \ 7.8 &  12.7 $\Rightarrow$ \ 0.1 & 46.4 $\Rightarrow$ \ 9.6 &  44.1 $\Rightarrow$  43.8 & 0.26 ( \ 0.59\%) &  19.12 (75.62\%) \\
    \bottomrule
  \end{tabular}}
  \label{ts_domain_detail}
\end{table*}

\begin{table*}[!t]
\renewcommand\arraystretch{1.6}
  \centering
    \caption{'Authorized/Unauthorized Drop' of target-specified CUPI-Domain, CUTI-Domain~[27] and NTL~[14] on the DomainNet~[29]. The result of CUTI-Domain~[27] and NTL~[14] is obtained by reproducing its source code for a fair comparison. The best performance is indicated by the numbers in bold. Statistical significance (p-value $<$ 0.05~[60],~[61]) is denoted with: $^{\ast}$(CUPI-Domain vs. CUTI-Domain~[27]) and $^{\diamond}$(CUPI-Domain vs. NTL~[14]).}
  \resizebox{1\textwidth}{!}{
  \begin{tabular}{c|ccc|ccc}
    \toprule
    \multirow{2}{*}{Domain / Drop} & \multicolumn{3}{c|}{Authorized Drop$\downarrow$} & \multicolumn{3}{c}{Unauthorized Drop$\uparrow$} \\
    \cline{2-7} 
    & CUPI-Domain & CUTI-Domain~[27] & NTL~[14] & CUPI-Domain & CUTI-Domain~[27] & NTL~[14] \\
    \midrule   
    clipart   & \bf 4.99 ( \ 7.42\%) & \ 7.69 (  11.45\%) & \ 7.78 (11.57\%) &  18.34 (71.44\%) &  15.32 (62.19\%) & \bf  20.99 (78.45\%)\\
    infograph & \bf 0.80 ( \ 3.48\%) & \ 4.39 (  19.08\%) & \ 7.65 (33.28\%) &  14.83 (79.79\%) &  12.76 (72.61\%) & \bf  15.49 (86.68\%)\\
    painting  & \bf 0.14 ( \ 0.49\%) & \ 0.81 ( \ 2.80\%) & \ 8.08 (27.94\%) &  11.74 (65.61\%) &  10.15 (60.67\%) & \bf  20.32 (89.05\%)\\
    quickdraw & \bf 6.95 (  13.62\%) &  10.70 (  20.97\%) &  48.12 (94.31\%) & \bf \ 9.23 (97.73\%) & \ 9.11 (96.15\%) & \ 8.74 (94.60\%)\\
    real      & \bf 8.15 (  10.44\%) & \ 8.25 (  10.57\%) &  13.00 (16.65\%) &  15.01 (66.65\%) &  13.86 (61.33\%) & \bf  17.30 (73.68\%)\\
    sketch    & \bf 0.26 ( \ 0.59\%) & \ 3.11 ( \ 7.04\%) &  11.74 (26.63\%) &  19.12 (75.62\%) &  17.72 (71.01\%) & \bf  23.35 (89.27\%)\\
    \midrule
    Mean & \bf \ 3.55 ( \ 6.01\%)$^{\ast \diamond}$ & 5.82 (  11.98\%) & 16.06 (35.06\%) & 14.71 (76.14\%)$^{\ast \diamond}$ & 13.15 (70.66\%) & \bf  17.70 (85.29\%)\\
    \bottomrule
  \end{tabular}}
  \label{ts_domain_compare}
\end{table*}

\begin{table*}[!t]
\renewcommand\arraystretch{1.6}
  \centering
  \caption{The results of ownership verification by SL and CUPI-Domain, CUTI-Domain~[27] and NTL~[14]. 'Training Methods' shows SL and CUPI-Domain accuracy on the on authorized domain without/with watermark. ‘Watermark Removal Approaches on CUPI’ represents CUPI-Domain accuracy after/before attacks by watermark removal methods: FTAL~[25], RTAL~[25], EWC~[26], AU~[26], and watermark overwriting followed by NTL~[14]. 'Average Drop' represents the average decrease in test dataset accuracy with a watermark compared to without, after applying watermark removal methods. 
  The result of CUTI-Domain~[27] and NTL~[14] is obtained by reproducing its source code for a fair comparison. The best performance is indicated by the numbers in bold. Statistical significance (p-value $<$ 0.05~[60],~[61]) is denoted with: $^{\ast}$(CUPI-Domain vs. CUTI-Domain~[27]) and $^{\diamond}$(CUPI-Domain vs. NTL~[14]).}
  \resizebox{1\textwidth}{!}{
  \begin{tabular}{c|cc|ccccc|ccc}
    \toprule
    \multirow{3}{*}{\makecell[c]{Source \\ without \\ Patch}} & \multicolumn{2}{c|}{Training Methods ($\%$)} & \multicolumn{5}{c|}{Watermark Removal Approaches on CUPI ($\%$)} & \multicolumn{3}{c}{Average Drop$\uparrow$}\\
    \cline{2-11}
    & \makecell[c]{SL} & CUPI & FTAL~[25] & RTAL~[25] & EWC~[26] & AU~[26] & Overwriting & \multirow{2}{*}{CUPI} &\multirow{2}{*}{CUTI~[27]} & \multirow{2}{*}{NTL~[14]}\\
    & \multicolumn{2}{c|}{[Test w/o Watermark ($\%$)]} & \multicolumn{5}{c|}{[Test w/o Watermark ($\%$)]}\\
    \midrule
    MNIST     & 99.0 $/$ 99.3 & \ 8.2 $/$ 99.5 & \ 5.5 $/$  100.0 & \ 7.4 $/$ \ 99.6 & \ 9.4 $/$  100.0 & \ 6.6 $/$  100.0 & \ 8.4 $/$ \ 99.4 & \bf 92.3 & 89.9 & 88.4 \\
    USPS      & 99.8 $/$ 99.8 & \ 6.5 $/$ 99.8 & \ 5.1 $/$  100.0 & \ 5.5 $/$  100.0 & \ 8.0 $/$  100.0 & \ 6.3 $/$  100.0 & \ 6.1 $/$ \ 99.7 & \bf 93.8 & 90.9 & 85.7 \\
    SVHN      & 91.3 $/$ 92.3 & \ 6.1 $/$ 90.0 & \ 7.0 $/$ \ 94.5 &  10.5 $/$ \ 93.4 & \ 9.4 $/$  100.0 & \ 6.6 $/$ \ 95.1 & \ 7.1 $/$ \ 89.8 & \bf 87.7 & \bf 87.7 & 79.0 \\
    MNIST-M   & 96.6 $/$ 96.0 &  14.7 $/$ 94.9 &  16.8 $/$ \ 97.7 &  21.9 $/$ \ 95.3 &  13.9 $/$  100.0 &  12.8 $/$ \ 96.1 &  15.9 $/$ \ 95.5 & \bf 81.5 & \bf 81.5 & 77.3 \\
    CIEAR10   & 83.3 $/$ 75.1 & \ 9.0 $/$ 85.0 & \ 9.4 $/$ \ 98.0 &  14.1 $/$ \ 96.9 &  14.6 $/$  100.0 & \ 8.3 $/$ \ 99.7 &  12.5 $/$ \ 95.7 & \bf 86.3 & 81.7 & 74.6 \\
    STL10     & 87.9 $/$ 93.2 &  12.3 $/$ 88.1 &  12.9 $/$ \ 98.4 &  18.0 $/$ \ 98.4 &  13.9 $/$  100.0 &  14.9 $/$ \ 94.8 &  15.6 $/$ \ 89.1 & \bf 81.1 & 74.0 & 74.0 \\
    VisDA     & 93.6 $/$ 92.2 & \ 8.6 $/$ 95.2 &  19.9 $/$  100.0 &  22.2 $/$ \ 98.6 &  14.9 $/$  100.0 & \ 9.0 $/$  100.0 &  21.5 $/$ \ 98.8 & \bf 82.0 & 78.0 & 76.8 \\
    Art       & 68.2 $/$ 79.2 & \ 1.0 $/$ 79.2 & \ 1.0 $/$  100.0 & \ 2.1 $/$  100.0 & \ 0.8 $/$  100.0 & \ 1.7 $/$  100.0 & \ 4.2 $/$  100.0 & \bf 98.0 & \bf 98.0 & 95.8 \\
    Clipart   & 75.3 $/$ 79.2 & \ 1.3 $/$ 77.6 & \ 1.0 $/$ \ 97.9 & \ 2.1 $/$ \ 97.9 & \ 2.5 $/$ \ 99.2 & \ 2.5 $/$ \ 98.3 & \ 5.2 $/$ \ 98.9 & \bf 95.8 & 93.5 & 91.8 \\
    Product   & 85.9 $/$ 92.4 & \ 0.8 $/$ 92.7 & \ 1.0 $/$  100.0 & \ 1.0 $/$  100.0 & \ 0.8 $/$  100.0 & \ 0.8 $/$  100.0 & \ 1.0 $/$  100.0 & \bf 99.0 & 94.2 & 95.4 \\
    RealWorld & 72.7 $/$ 82.8 & \ 0.3 $/$ 82.8 & \ 2.1 $/$  100.0 & \ 2.1 $/$  100.0 & \ 2.5 $/$  100.0 & \ 3.3 $/$  100.0 & \ 3.1 $/$  100.0 & \bf 97.4 & 92.3 & 93.5 \\
    clipart   & 53.9 $/$ 59.8 & 0.9 $/$ 61.1 & 5.9 $/$ 78.4 & \ 6.7 $/$ 76.7 & 1.4 $/$ 79.2 & 1.4 $/$ 79.1 & 6.7 $/$ 78.7 & \bf 74.0 & 63.9 & 61.1 \\
    infograph & 13.2 $/$ 18.2 & 0.3 $/$ 18.6 & 1.3 $/$ 39.4 & \ 1.3 $/$ 32.3 & 0.3 $/$ 42.7 & 0.3 $/$ 39.8 & 1.4 $/$ 39.7 & \bf 37.8 & 25.5 & 20.6 \\
    painting  & 35.8 $/$ 45.7 & 0.4 $/$ 39.5 & 3.5 $/$ 62.9 & \ 4.4 $/$ 58.8 & 0.7 $/$ 59.0 & 0.7 $/$ 59.2 & 1.9 $/$ 63.2 & \bf 58.4 & 56.4 & 53.3 \\
    quickdraw & 41.8 $/$ 42.0 & 0.0 $/$ 40.9 & 0.2 $/$ 57.6 & \ 2.0 $/$ 53.1 & 0.1 $/$ 55.6 & 0.1 $/$ 54.3 & 0.3 $/$ 58.1 & \bf 55.2 & 51.9 & 33.4 \\
    real      & 52.6 $/$ 67.3 & 0.7 $/$ 63.8 & 3.9 $/$ 79.0 & \ 6.2 $/$ 75.7 & 1.4 $/$ 78.1 & 0.8 $/$ 78.3 & 4.9 $/$ 79.7 & \bf 74.7 & 73.7 & 73.2 \\
    sketch    & 36.6 $/$ 39.3 & 0.2 $/$ 44.5 & 4.6 $/$ 64.5 &  13.5 $/$ 59.0 & 0.7 $/$ 66.7 & 0.2 $/$ 66.8 & 4.1 $/$ 64.3 & \bf 59.6 & 41.7 & 42.1 \\
    \midrule
    Mean & / & / & / & / & / & / & / & \bf 79.7$^{\ast \diamond}$ & 75.0 & 71.5 \\
    \bottomrule
  \end{tabular}}
  \label{ov_detail}
\end{table*}

\begin{table*}[!t]
\renewcommand\arraystretch{1.6}
  \centering
    \caption{The accuracy ($\%$) of target-free CUPI-Domain on digit datasets. The vertical/horizontal axis denotes the authorized/unauthorized domain. In each task, the left of '\(\Rightarrow\)' shows the test accuracy of SL on the unauthorized domain, while the right side presents the accuracy of CUPI-Domain. 'Authorized/Unauthorized Drop' indicate the drop (relative drop) of CUPI-Domain relative to SL on authorized/unauthorized domains.}
  \resizebox{1\textwidth}{!}{
  \begin{tabular}{c|cccc|cc}
    \toprule
    Authorized/Unauthorized &  MNIST & USPS & SVHN & MNIST-M & Authorized Drop$\downarrow$ & Unauthorized Drop$\uparrow$ \\
    \midrule    
    MNIST~[52]   & 99.2 $\Rightarrow$  99.2 & 98.0 $\Rightarrow$ \ 6.7 & 38.2 $\Rightarrow$ \ 6.5 & 67.8 $\Rightarrow$ \ 9.6 & 0.03 (0.03\%) & 60.43 (87.39\%) \\ 
    USPS~[53]    & 92.6 $\Rightarrow$ \ 8.9 & 99.7 $\Rightarrow$  99.4 & 25.5 $\Rightarrow$ \ 6.5 & 41.2 $\Rightarrow$ \ 8.1 & 0.33 (0.33\%) & 45.29 (81.80\%) \\
    SVHN~[54]    & 66.7 $\Rightarrow$  11.8 & 70.5 $\Rightarrow$  14.1 & 91.2 $\Rightarrow$  90.1 & 34.6 $\Rightarrow$  11.4 & 1.10 (1.20\%) & 44.87 (76.53\%) \\
    MNIST-M~[55] & 98.4 $\Rightarrow$ \ 9.9 & 88.4 $\Rightarrow$ \ 7.6 & 46.3 $\Rightarrow$ \ 6.1 & 95.4 $\Rightarrow$  95.2 & 0.19 (0.20\%) & 69.82 (89.36\%) \\   
    \bottomrule
  \end{tabular}}
  \label{tf_dight_detail}
\end{table*}

\begin{table*}[!t]
\renewcommand\arraystretch{1.6}
  \centering
    \caption{The accuracy ($\%$) of target-free CUPI-Domain on the Office-Home-65~[28]. The vertical/horizontal axis denotes the authorized/unauthorized domain. In each task, the left of '\(\Rightarrow\)' shows the test accuracy of SL on the unauthorized domain, while the right side presents the accuracy of CUPI-Domain. 'Authorized/Unauthorized Drop' indicate the drop (relative drop) of CUPI-Domain relative to SL on authorized/unauthorized domains.}
  \resizebox{1\textwidth}{!}{
  \begin{tabular}{c|cccc|cc}
    \toprule
    Authorized/Unauthorized & Artistic & Clipart & Product & RealWorld & Authorized Drop$\downarrow$ & Unauthorized Drop$\uparrow$ \\
    \midrule   
    Artistic  & 76.3 $\Rightarrow$  75.0 & 47.1 $\Rightarrow$  31.0 & 64.9 $\Rightarrow$ 48.7 & 72.2 $\Rightarrow$ 62.8  & 1.30 (1.70\%) & 13.92 (24.08\%)  \\ 
    Clipart   & 57.8 $\Rightarrow$ \ 8.1 & 80.1 $\Rightarrow$  73.7 & 63.5 $\Rightarrow$ 16.4 & 68.8 $\Rightarrow$ 14.3  & 6.40 (7.99\%) & 50.41 (79.78\%)  \\ 
    Product   & 56.6 $\Rightarrow$  13.8 & 45.2 $\Rightarrow$ \ 4.4 & 92.7 $\Rightarrow$ 86.2 & 72.7 $\Rightarrow$ 35.7  & 6.50 (7.01\%) & 40.18 (72.24\%)  \\ 
    RealWorld & 63.8 $\Rightarrow$  54.7 & 49.2 $\Rightarrow$  22.7 & 75.5 $\Rightarrow$ 70.8 & 84.4 $\Rightarrow$ 80.2  & 4.19 (4.97\%) & 13.44 (24.80\%)  \\
    \bottomrule
  \end{tabular}}
  \label{tf_home_detail}
\end{table*}

\begin{table*}[!t]
\renewcommand\arraystretch{1.6}
  \centering
    \caption{'Authorized/Unauthorized Drop' of target-free CUPI-Domain, CUTI-Domain~[27] and NTL~[14] on the Office-Home-65~[28]. The result of CUTI-Domain~[27] and NTL~[14] is obtained by reproducing its source code for a fair comparison. The best performance is indicated by the numbers in bold. Statistical significance (p-value $<$ 0.05~[60],~[61]) is denoted with: $^{\ast}$(CUPI-Domain vs. CUTI-Domain~[27]) and $^{\diamond}$(CUPI-Domain vs. NTL~[14]).}
  \resizebox{1\textwidth}{!}{
  \begin{tabular}{c|ccc|ccc}
    \toprule
    \multirow{2}{*}{Domain / Drop} & \multicolumn{3}{c|}{Authorized Drop$\downarrow$} & \multicolumn{3}{c}{Unauthorized Drop$\uparrow$} \\
    \cline{2-7} 
     & CUPI-Domain & CUTI-Domain~[27] & NTL~[14] & CUPI-Domain & CUTI-Domain~[27] & NTL~[14] \\
    \midrule   
    Artistic  & \bf 1.30 (1.70\%) & 3.90 (5.12\%) & \bf 1.30 (1.70\%) & 13.92 (24.08\%) & \bf 14.09 (23.98\%) & 4.54 (7.32\%)  \\ 
    Clipart   & \bf 6.40 (7.99\%) & 7.97 (9.94\%) & 6.92 (8.64\%) & \bf 50.41 (79.78\%) & 48.50 (76.77\%) & 6.49 (10.05\%) \\ 
    Product   & \bf 6.50 (7.01\%) & 7.28 (7.86\%) & \bf 6.50 (7.01\%) & \bf 40.18 (72.24\%) & 35.14 (63.96\%) & 5.19 (9.35\%)  \\ 
    RealWorld & \bf 4.19 (4.97\%) & 4.97 (5.89\%) & \bf 4.19 (4.97\%) & \bf 13.44 (24.80\%) & 11.18 (19.82\%) & 5.28 (9.64\%)  \\       
    \midrule
    Mean      & \bf 4.60 (5.42\%)$^{\ast}$ & 6.03 (7.20\%) & 4.91 (5.58\%) & \bf 29.49 (50.23\%)$^{\ast \diamond}$ & 27.23 (46.13\%) & 5.38 (9.09\%)  \\ 
    \bottomrule
  \end{tabular}}
  \label{tf_home_compare}
\end{table*}

\begin{table*}[!t]
\renewcommand\arraystretch{1.6}
  \centering
    \caption{The accuracy ($\%$) of target-free CUPI-Domain on the DomainNet~[29]. The vertical/horizontal axis denotes the authorized/unauthorized domain. In each task, the left of '\(\Rightarrow\)' shows the test accuracy of SL on the unauthorized domain, while the right side presents the accuracy of CUPI-Domain. 'Authorized/Unauthorized Drop' indicate the drop (relative drop) of CUPI-Domain relative to SL on authorized/unauthorized domains.}
  \resizebox{1\textwidth}{!}{
  \begin{tabular}{c|cccccc|cc}
    \toprule
    Authorized/Unauthorized & clipart & infograph & painting & quickdraw & real & sketch & Authorized Drop$\downarrow$ & Unauthorized Drop$\uparrow$ \\
    \midrule   
    clipart   & 67.2 $\Rightarrow$  62.3 &  12.8 $\Rightarrow$ \ 8.7 &  29.7 $\Rightarrow$  25.6 &  13.8 $\Rightarrow$ \ 0.3 & 51.2 $\Rightarrow$  41.5 &  31.6 $\Rightarrow$  31.6 & \ 4.91 ( \ 7.30\%) & \ 6.31 (32.61\%) \\ 
    infograph & 21.2 $\Rightarrow$ \ 9.4 &  23.0 $\Rightarrow$  16.2 &  21.1 $\Rightarrow$  12.7 & \ 3.6 $\Rightarrow$ \ 0.3 & 35.1 $\Rightarrow$  17.9 &  16.2 $\Rightarrow$ \ 8.5 & \ 6.75 (  29.36\%) & \ 9.69 (56.64\%) \\
    painting  & 28.9 $\Rightarrow$  25.0 &  13.6 $\Rightarrow$  11.7 &  50.0 $\Rightarrow$  47.9 & \ 3.7 $\Rightarrow$ \ 0.2 & 45.6 $\Rightarrow$  41.4 &  22.1 $\Rightarrow$  20.1 & \ 2.10 ( \ 7.27\%) & \ 3.13 (27.96\%) \\ 
    quickdraw & 17.2 $\Rightarrow$ \ 0.1 & \ 2.3 $\Rightarrow$ \ 0.1 & \ 6.1 $\Rightarrow$ \ 0.0 &  51.0 $\Rightarrow$  43.3 & 12.6 $\Rightarrow$ \ 0.1 & \ 8.6 $\Rightarrow$ \ 0.1 & \ 7.67 (  15.04\%) & \ 9.29 (98.66\%) \\
    real      & 39.6 $\Rightarrow$  26.7 &  13.4 $\Rightarrow$  11.4 &  40.3 $\Rightarrow$  32.5 & \ 6.4 $\Rightarrow$ \ 0.2 & 78.1 $\Rightarrow$  66.4 &  25.8 $\Rightarrow$  14.3 &  11.70 (  14.99\%) & \ 8.10 (41.87\%) \\
    sketch    & 35.8 $\Rightarrow$  25.8 &  11.3 $\Rightarrow$ \ 5.6 &  25.7 $\Rightarrow$  18.3 &  12.7 $\Rightarrow$ \ 0.8 & 46.4 $\Rightarrow$  27.2 &  44.1 $\Rightarrow$  43.9 & \ 0.22 ( \ 0.50\%) &  10.81 (48.33\%) \\

    \bottomrule
  \end{tabular}}
  \label{tf_domain_detail}
\end{table*}

\begin{table*}[!t]
\renewcommand\arraystretch{1.6}
  \centering
    \caption{'Authorized/Unauthorized Drop' of target-free CUPI-Domain, CUTI-Domain~[27] and NTL~[14] on the DomainNet~[29]. The result of CUTI-Domain~[27] and NTL~[14] is obtained by reproducing its source code for a fair comparison. 
    The best performance is indicated by the numbers in bold. Statistical significance (p-value $<$ 0.05~[60],~[61]) is denoted with: $^{\ast}$(CUPI-Domain vs. CUTI-Domain~[27]) and $^{\diamond}$(CUPI-Domain vs. NTL~[14]).}
  \resizebox{1\textwidth}{!}{
  \begin{tabular}{c|ccc|ccc}
    \toprule
    \multirow{2}{*}{Domain / Drop} & \multicolumn{3}{c|}{Authorized Drop$\downarrow$} & \multicolumn{3}{c}{Unauthorized Drop$\uparrow$} \\
    \cline{2-7} 
    & CUPI-Domain & CUTI-Domain~[27] & NTL~[14] & CUPI-Domain & CUTI-Domain~[27] & NTL~[14] \\
    \midrule   
    clipart   & \bf \ 4.91 (\ 7.30\%) &  10.96 (  16.30\%) & \ 8.15 (  12.13\%) & \ 6.31 (32.61\%) & \bf \ 7.72 (36.92\%) & 1.46 (\ 4.44\%)  \\ 
    infograph & \ 6.75 ( 29.36\%) & \ 4.87 (  21.17\%) & \bf \ 4.17 (  18.12\%) & \bf \ 9.69 (56.64\%) & \ 7.85 (47.40\%) & 5.32 ( 32.92\%)  \\ 
    painting  & \ 2.10 (\ 7.27\%) & \ 0.81 ( \ 2.80\%) & \bf \ 0.26 ( \ 0.90\%) & \bf \ 3.13 (27.96\%) & \ 1.74 (20.78\%) & 0.74 (\ 7.21\%)  \\ 
    quickdraw & \bf \ 7.67 ( 15.04\%) &  19.85 (  38.91\%) &  25.00 (  49.00\%) & \bf \ 9.29 (98.66\%) & \ 9.28 (99.02\%) & 8.66 ( 95.26\%)  \\ 
    real      &  11.70 ( 14.99\%) &  12.32 (  15.78\%) & \bf \ 7.59 ( \ 9.73\%) & \bf \ 8.10 (41.87\%) & \ 5.64 (31.50\%) & 1.38 ( 13.34\%)  \\ 
    sketch    & \bf \ 0.22 (\ 0.50\%) & \ 1.68 ( \ 3.82\%) & \ 8.37 (  18.99\%) &  \bf 10.81 (48.33\%) &  10.80 (49.79\%) & 8.66 ( 32.39\%)  \\ 
    \midrule
    Mean      & \bf \ 5.56 ( 12.41\%)$^{\ast}$ & \ 8.42 (  16.46\%) & \ 8.92 (  18.14\%) & \bf \ 7.89 (51.01\%)$^{\ast \diamond}$ & \ 7.17 (47.57\%) & 4.37 ( 30.93\%)  \\       
    \bottomrule
  \end{tabular}}
  \label{tf_domain_compare}
\end{table*}

\begin{table*}
  \renewcommand\arraystretch{1.6}
  \centering
  \caption{'Authorized Domain', 'Other Domains' and 'Drop' of applicability authorization scheme 1 of CUPI-Domain, CUTI-Domain~[27] and NTL~[14] on digit datasets, CIFAR10\&STL10 and VisDA-2017. The result of CUTI-Domain~[27] and NTL~[14] is obtained by reproducing its source code for a fair comparison. The best performance is indicated by the numbers in bold. Statistical significance (p-value $<$ 0.05~[60],~[61]) is denoted with: $^{\ast}$(CUPI-Domain vs. CUTI-Domain~[27]) and $^{\diamond}$(CUPI-Domain vs. NTL~[14]).}
  \resizebox{1.\textwidth}{!}{
  
\begin{tabular}{c|ccc|ccc|ccc}
\toprule
\multirow{2}{*}{Domain / Accuracy} & \multicolumn{3}{c|}{Authorized Domain$\uparrow$} & \multicolumn{3}{c|}{Other Domains$\downarrow$} & \multicolumn{3}{c}{Drop$\uparrow$} \\
\cline{2-10}
& CUPI & CUTI~[27] & NTL~[14] & CUPI & CUTI~[27] & NTL~[14] & CUPI & CUTI~[27] & NTL~[14] \\
\midrule
MNIST~[52]    & \bf  100.00 & \bf 100.00 &      99.80 & \bf \ 8.81 & 13.73 & 14.49 & \bf 91.19(91.19\%) & 86.27(86.27\%) & 85.31(85.49\%) \\ 
USPS~[53]     & \bf \ 99.70 &    \ 99.20 &      98.50 & \bf \ 8.57 & 10.19 & 13.30 & \bf 91.13(91.40\%) & 89.01(89.73\%) & 85.20(86.50\%) \\
SVHN~[54]     & \bf \ 99.60 &    \ 99.10 &      99.30 & \bf  10.34 & 13.24 & 15.79 & \bf 89.26(89.62\%) & 85.86(86.64\%) & 83.51(84.10\%) \\
MNIST-M~[55]  & \bf \ 99.90 &    \ 99.50 &      99.50 & \bf  10.24 & 12.84 & 14.01 & \bf 89.66(89.75\%) & 86.66(87.09\%) & 85.49(85.92\%) \\
CIFAR10~[56]  & \bf  100.00 &      97.90 &      97.50 & \bf  10.80 & 22.40 & 24.60 & \bf 89.20(89.20\%) & 75.50(77.12\%) & 72.90(74.77\%) \\
STL10~[57]    & \bf \ 99.69 &      99.90 &      98.60 & \bf  17.99 & 19.17 & 20.50 & \bf 81.70(81.96\%) & 80.73(80.81\%) & 78.10(79.21\%) \\        
VisDA-T~[59]  & \bf  100.00 & \bf 100.00 & \bf 100.00 & \bf  12.15 & 17.77 & 23.30 & \bf 87.85(87.85\%) & 82.23(82.23\%) & 76.70(76.70\%) \\ 
\midrule
Mean     & \bf \ 99.84$^{\diamond}$  &   \ 99.37 &      99.03 & \bf  11.27$^{\ast \diamond}$ & 15.62 & 18.00 & \bf 88.57(88.71\%)$^{\ast \diamond}$ & 83.75(84.27\%) & 81.03(81.81\%) \\  
\bottomrule
\end{tabular}
}
  \vskip -10pt
  \label{aa_1_compare}
\end{table*}

\begin{table*}
  \renewcommand\arraystretch{1.6}
  \centering
  \caption{The accuracy ($\%$) of applicability authorization scheme 1 of CUPI-Domain on digit datasets, CIFAR10\&STL10 and VisDA-2017. The vertical axis represents the name of authorized domain, while the horizontal axis represents the name of test domain with/without patch. 'Authorized/Other Domains' represents the average accuracy on the authorized/other domains, and 'Drop' denote the drop (relative drop) from authorized to other domains.}
  \resizebox{1.\textwidth}{!}{
  
\begin{tabular}{c|cccc|cccc|ccc}
\toprule
\multirow{3}{*}{\begin{tabular}[c]{@{}c@{}}AUT \\ with \\ Path\end{tabular}} & \multicolumn{4}{c|}{\multirow{2}{*}{Test Domain with Patch (\%)}} & \multicolumn{4}{c|}{\multirow{2}{*}{Test Domain without Patch (\%)}} & \multicolumn{1}{c}{\multirow{3}{*}{\begin{tabular}[c]{@{}c@{}}Authorized \\ Domain with \\ Patch$\uparrow$\end{tabular}}} & \multicolumn{1}{c}{\multirow{3}{*}{\begin{tabular}[c]{@{}c@{}} Other \\ Domains \\ $\downarrow$\end{tabular}}} & \multicolumn{1}{c}{\multirow{3}{*}{\begin{tabular}[c]{@{}c@{}} Drop \\ (Relative Drop) \\ $\uparrow$\end{tabular}}} \\
 & \multicolumn{4}{c|}{} & \multicolumn{4}{c|}{} \\ 
\cline{2-9}
 & MNIST~[52] & USPS & SVHN & \multicolumn{1}{c|}{MNIST-M} & MNIST & USPS & SVHN & MNIST-M & \multicolumn{1}{c}{} & \multicolumn{1}{c}{} \\ 
\midrule
MNIST~[52]   &  100.0 & \ 9.6 & \ 7.2 & \multicolumn{1}{c|}{ \ 8.9} &  11.0 & 9.6 & \ 6.8 & \ 8.6 &  100.00 & \ 8.81 & 91.19(91.19\%) \\ 
USPS~[53]    &  \ 6.6 &  99.7 & \ 9.3 & \multicolumn{1}{c|}{ \ 9.3} & \ 8.9 & 7.2 & \ 9.4 & \ 9.3 & \ 99.70 & \ 8.57 & 91.13(91.40\%) \\
SVHN~[54]    & \ 10.2 & \ 9.0 &  99.6 & \multicolumn{1}{c|}{  10.6} &  10.8 & 9.0 &  12.0 &  10.8 & \ 99.60 &  10.34 & 89.26(89.62\%) \\
MNIST-M~[55] &  \ 8.5 & \ 7.6 &  20.2 & \multicolumn{1}{c|}{  99.9} & \ 8.4 & 7.5 & \ 9.6 & \ 9.9 & \ 99.90 &  10.24 & 89.66(89.75\%) \\
\midrule
 & \multicolumn{2}{c}{CIFAR} & \multicolumn{2}{c|}{STL} & \multicolumn{2}{c}{CIFAR} & \multicolumn{2}{c|}{STL} & \multicolumn{3}{c}{/} \\ 
\midrule
CIFAR10~[56]   & \multicolumn{2}{c}{  100.0} & \multicolumn{2}{c|}{12.2} & \multicolumn{2}{c}{\ 9.5} & \multicolumn{2}{c|}{10.7} & 100.00 & 10.80 & 89.20(89.20\%) \\
STL10~[57]     & \multicolumn{2}{c}{ \ 21.4} & \multicolumn{2}{c|}{99.7} & \multicolumn{2}{c}{ 14.6} & \multicolumn{2}{c|}{18.0} &  99.69 & 17.99 & 81.70(81.96\%) \\         
\midrule
 & \multicolumn{2}{c}{VisDA-T} & \multicolumn{2}{c|}{VisDA-V} & \multicolumn{2}{c}{VisDA-T} & \multicolumn{2}{c|}{VisDA-V} & \multicolumn{3}{c}{/} \\ 
\midrule
VisDA-T~[59] & \multicolumn{2}{c}{100.0} & \multicolumn{2}{c|}{21.8} & \multicolumn{2}{c}{7.6} & \multicolumn{2}{c|}{7.1} & 100.00 & 12.15 & 87.85(87.85\%) \\ 
\bottomrule
\end{tabular}
}
  \vskip -10pt
  \label{aa_1_detail}
\end{table*}

\begin{table*}
  \renewcommand\arraystretch{1.6}
  \centering
  \caption{'Authorized Domain', 'Other Domains' and 'Drop' of applicability authorization scheme 2 of CUPI-Domain, CUTI-Domain~[27] and NTL~[14] on digit datasets, CIFAR10\&STL10 and VisDA-2017. The result of CUTI-Domain~[27] and NTL~[14] is obtained by reproducing its source code for a fair comparison. The best performance is indicated by the numbers in bold. Statistical significance (p-value $<$ 0.05~[60],~[61]) is denoted with: $^{\ast}$(CUPI-Domain vs. CUTI-Domain~[27]) and $^{\diamond}$(CUPI-Domain vs. NTL~[14]).}
  \resizebox{1.\textwidth}{!}{
  
\begin{tabular}{c|ccc|ccc|ccc}
\toprule
\multirow{2}{*}{Domain / Accuracy} & \multicolumn{3}{c|}{Authorized Domain$\uparrow$} & \multicolumn{3}{c|}{Other Domains$\downarrow$} & \multicolumn{3}{c}{Drop$\uparrow$} \\
\cline{2-10}
& CUPI & CUTI~[27] & NTL~[14] & CUPI & CUTI~[27] & NTL~[14] & CUPI & CUTI~[27] & NTL~[14] \\
\midrule
MNIST~[52]    & \bf  100.00 &      99.90 &      99.50 & \bf \ 8.74 & 15.52 & 16.56 & \bf 91.27(91.27\%) & 84.38(84.46\%) & 82.94(83.36\%) \\ 
USPS~[53]     & \bf \ 99.69 &    \ 99.27 &      99.50 & \bf \ 9.24 &  9.51 & 10.73 & \bf 90.45(90.73\%) & 89.76(90.42\%) & 88.77(89.22\%) \\
SVHN~[54]     & \bf \ 99.69 & \bf  99.69 &      99.19 & \bf  14.29 & 19.91 & 22.88 & \bf 85.40(85.67\%) & 79.78(80.03\%) & 76.31(76.93\%) \\
MMNIST-M~[55] & \bf \ 99.06 &    \ 98.29 &      98.89 & \bf  18.05 & 18.71 & 21.03 & \bf 81.01(81.78\%) & 79.58(80.97\%) & 77.87(78.74\%) \\
CIFAR10~[56]  & \bf  100.00 & \bf 100.00 & \bf 100.00 & \bf  10.97 & 12.40 & 15.24 & \bf 89.03(89.03\%) & 87.60(87.60\%) & 84.76(84.76\%) \\
STL10~[57]    & \bf \ 99.79 &      99.17 & \bf  99.79 & \bf  25.35 & 25.90 & 29.31 & \bf 74.45(74.60\%) & 73.26(73.88\%) & 70.49(70.63\%) \\        
VisDA-T~[59]  & \bf  100.00 & \bf 100.00 &      98.54 & \bf  13.13 & 14.44 & 14.44 & \bf 86.88(86.88\%) & 85.56(85.56\%) & 84.10(85.34\%) \\ 
\midrule
Mean     & \bf \ 99.75$^{\ast \diamond}$ &    \ 99.47 &      99.34 & \bf  14.25$^{\ast \diamond}$ & 16.63 & 18.60 & \bf 85.50(85.71\%)$^{\ast \diamond}$ & 85.85(83.27\%) & 80.75(81.28\%) \\  
\bottomrule
\end{tabular}
}
  \vskip -10pt
  \label{aa_2_compare}
\end{table*}

\begin{table*}
  \renewcommand\arraystretch{1.6}
  \centering
  \caption{The accuracy ($\%$) of applicability authorization scheme 2 of CUPI-Domain on digit datasets, CIFAR10\&STL10 and VisDA-2017. The vertical axis represents the name of authorized domain, while the horizontal axis represents the name of test domain with/without patch.'Authorized/Other Domains' represents the average accuracy on the authorized/other domains, and 'Drop' denote the drop (relative drop) from authorized to other domains.}
  \resizebox{1.\textwidth}{!}{
  
\begin{tabular}{c|cccc|cccc|ccc}
\toprule
\multirow{3}{*}{\begin{tabular}[c]{@{}c@{}}AUT \\ without \\ Path\end{tabular}} & \multicolumn{4}{c|}{\multirow{2}{*}{Test Domain with Patch (\%)}} & \multicolumn{4}{c|}{\multirow{2}{*}{Test Domain without Patch (\%)}} & \multicolumn{1}{c}{\multirow{3}{*}{\begin{tabular}[c]{@{}c@{}}Authorized \\ Domain without \\ Patch$\uparrow$\end{tabular}}} & \multicolumn{1}{c}{\multirow{3}{*}{\begin{tabular}[c]{@{}c@{}} Other \\ Domains \\ $\downarrow$\end{tabular}}} & \multicolumn{1}{c}{\multirow{3}{*}{\begin{tabular}[c]{@{}c@{}} Drop \\ (Relative Drop) \\ $\uparrow$\end{tabular}}} \\
 & \multicolumn{4}{c|}{} & \multicolumn{4}{c|}{} \\ 
\cline{2-9}
 & MNIST~[52] & USPS & SVHN & \multicolumn{1}{c|}{MNIST-M} & MNIST & USPS & SVHN & MNIST-M & \multicolumn{1}{c}{} & \multicolumn{1}{c}{} \\ 
\midrule
MNIST~[52]   & \ 9.6 & \ 9.4 & \ 7.0 & \multicolumn{1}{c|}{ \ 8.4} &  100.0 & \ 9.1 & \ 7.1 &  10.6 &  100.00 & \ 8.74 & 91.27(91.27\%) \\ 
USPS~[53]    & \ 9.9 & \ 9.5 & \ 6.8 & \multicolumn{1}{c|}{ \ 8.2} & \ 15.1 &  99.7 & \ 7.0 & \ 8.2 & \ 99.69 & \ 9.24 & 90.45(90.73\%) \\
SVHN~[54]    &  10.9 &  14.6 &  17.9 & \multicolumn{1}{c|}{  10.9} & \ 10.5 &  19.4 &  99.7 &  15.7 & \ 99.69 &  14.29 & 85.40(85.67\%) \\
MNIST-M~[55] &  10.8 &  14.3 &  18.1 & \multicolumn{1}{c|}{  13.4} & \ 28.3 &  14.2 &  27.2 &  99.1 & \ 99.06 &  18.05 & 81.01(81.78\%) \\
\midrule
 & \multicolumn{2}{c}{CIFAR} & \multicolumn{2}{c|}{STL} & \multicolumn{2}{c}{CIFAR} & \multicolumn{2}{c|}{STL} & \multicolumn{3}{c}{/} \\ 
\midrule
CIFAR10~[56]   & \multicolumn{2}{c}{11.7} & \multicolumn{2}{c|}{ 10.6} & \multicolumn{2}{c}{ 100.0} & \multicolumn{2}{c|}{10.6} & 100.00 & 10.97 & 89.03(89.03\%) \\
STL10~[57]     & \multicolumn{2}{c}{20.3} & \multicolumn{2}{c|}{\ 9.1} & \multicolumn{2}{c}{\ 46.7} & \multicolumn{2}{c|}{99.8} &   99.79 & 25.35 & 74.45(74.60\%) \\         
\midrule
 & \multicolumn{2}{c}{VisDA-T} & \multicolumn{2}{c|}{VisDA-V} & \multicolumn{2}{c}{VisDA-T} & \multicolumn{2}{c|}{VisDA-V} & \multicolumn{3}{c}{/} \\ 
\midrule
VisDA-T~[59] & \multicolumn{2}{c}{7.4} & \multicolumn{2}{c|}{7.4} & \multicolumn{2}{c}{100.0} & \multicolumn{2}{c|}{24.6} & 100.00 & 13.13 & 86.88(86.88\%) \\ 
\bottomrule
\end{tabular}
}
  \vskip -10pt
  \label{aa_2_detail}
\end{table*}

\begin{table*}
  \renewcommand\arraystretch{1.6}
  \centering
  \caption{'Authorized Domain', 'Other Domains' and 'Drop' of applicability authorization scheme 3 of CUPI-Domain, CUTI-Domain~[27] and NTL~[14] on digit datasets, CIFAR10\&STL10 and VisDA-2017. The result of CUTI-Domain~[27] and NTL~[14] is obtained by reproducing its source code for a fair comparison. The best performance is indicated by the numbers in bold. Statistical significance (p-value $<$ 0.05~[60],~[61]) is denoted with: $^{\ast}$(CUPI-Domain vs. CUTI-Domain~[27]) and $^{\diamond}$(CUPI-Domain vs. NTL~[14]).}
  \resizebox{1.\textwidth}{!}{
  
\begin{tabular}{c|ccc|ccc|ccc}
\toprule
\multirow{2}{*}{Domain / Accuracy} & \multicolumn{3}{c|}{Authorized Domain$\uparrow$} & \multicolumn{3}{c|}{Other Domains$\downarrow$} & \multicolumn{3}{c}{Drop$\uparrow$} \\
\cline{2-10}
& CUPI & CUTI~[27] & NTL~[14] & CUPI & CUTI~[27] & NTL~[14] & CUPI & CUTI~[27] & NTL~[14] \\
\midrule
MNIST~[52]    & \bf 99.60 & 99.43 & 99.31 & \bf \ 8.48 & 14.56 & 15.71 & \bf 91.12 (91.48\%) & 84.87 (85.36\%) & 83.60 (84.18\%) \\ 
USPS~[53]     & \bf 98.45 & 97.86 & 97.90 & \bf \ 9.25 &  9.76 & 12.10 & \bf 89.20 (90.60\%) & 88.10 (90.03\%) & 85.80 (87.64\%) \\
SVHN~[54]     & \bf 99.30 & 99.10 & 99.10 & \bf  13.50 & 15.63 & 18.23 & \bf 85.80 (86.40\%) & 83.47 (84.23\%) & 80.87 (81.60\%) \\
MMNIST-M~[55] & \bf 98.55 & 98.50 & 98.43 & \bf  18.65 & 19.24 & 19.28 & \bf 79.90 (81.08\%) & 79.26 (81.10\%) & 79.15 (80.41\%) \\
CIFAR10~[56]  & \bf 98.95 & 98.30 & 97.59 & \bf  10.15 & 15.47 & 18.17 & \bf 88.80 (89.74\%) & 82.83 (84.26\%) & 79.42 (81.38\%) \\
STL10~[57]    & \bf 98.85 & 98.26 & 98.10 & \bf  16.90 & 18.57 & 21.52 & \bf 81.95 (82.90\%) & 79.69 (81.10\%) & 76.58 (78.06\%) \\        
VisDA-T~[59]  & \bf 99.85 & 98.95 & 99.03 & \bf  12.10 & 15.20 & 15.96 & \bf 87.75 (87.88\%) & 83.75 (84.64\%) & 83.07 (83.88\%) \\ 
\midrule
Mean     & \bf \ 99.08$^{\ast \diamond}$ &    \ 98.77 &      98.64 & \bf  12.72$^{\ast \diamond}$ & 14.78 & 16.90 & \bf 86.36 (87.16\%)$^{\ast \diamond}$ & 84.00 (85.05\%) & 81.74 (82.88\%) \\  
\bottomrule
\end{tabular}
}
  \vskip -10pt
  \label{aa_3_compare}
\end{table*}

\begin{table*}
  \renewcommand\arraystretch{1.6}
  \centering
  \caption{The accuracy ($\%$) of applicability authorization scheme 3 of CUPI-Domain on digit datasets, CIFAR10\&STL10 and VisDA-2017. The vertical axis represents the name of authorized domain, while the horizontal axis represents the name of test domain with/without patch.'Authorized/Other Domains' represents the average accuracy on the authorized/other domains, and 'Drop' denote the drop (relative drop) from authorized to other domains.}  \resizebox{1.\textwidth}{!}{
  
\begin{tabular}{c|cccc|cccc|ccc}
\toprule
\multirow{3}{*}{\begin{tabular}[c]{@{}c@{}}AUT \\ without \\ Path\end{tabular}} & \multicolumn{4}{c|}{\multirow{2}{*}{Test Domain with Patch (\%)}} & \multicolumn{4}{c|}{\multirow{2}{*}{Test Domain without Patch (\%)}} & \multicolumn{1}{c}{\multirow{3}{*}{\begin{tabular}[c]{@{}c@{}}Authorized \\ Domain without \\ Patch$\uparrow$\end{tabular}}} & \multicolumn{1}{c}{\multirow{3}{*}{\begin{tabular}[c]{@{}c@{}} Other \\ Domains \\ $\downarrow$\end{tabular}}} & \multicolumn{1}{c}{\multirow{3}{*}{\begin{tabular}[c]{@{}c@{}} Drop \\ (Relative Drop) \\ $\uparrow$\end{tabular}}} \\
 & \multicolumn{4}{c|}{} & \multicolumn{4}{c|}{} \\ 
\cline{2-9}
 & MNIST~[52] & USPS & SVHN & \multicolumn{1}{c|}{MNIST-M} & MNIST & USPS & SVHN & MNIST-M & \multicolumn{1}{c}{} & \multicolumn{1}{c}{} \\ 
\midrule
MNIST~[52]   &  99.2 & \ 9.7 & \ 6.7 & \multicolumn{1}{c|}{ \ 8.6} &  100.0 & \ 9.4 & \ 6.8 & \ 9.7 & 99.60 & \ 8.48 & 91.12 (91.48\%) \\ 
USPS~[53]    & \ 9.2 &  99.2 & \ 6.8 & \multicolumn{1}{c|}{ \ 8.5} & \ 15.8 &  97.7 & \ 6.7 & \ 8.5 & 98.45 & \ 9.25 & 89.20 (90.60\%) \\
SVHN~[54]    &  11.4 &  14.0 &  98.9 & \multicolumn{1}{c|}{  10.4} & \ 12.3 &  17.8 &  99.7 &  15.1 & 99.30 &  13.50 & 85.80 (86.40\%) \\
MNIST-M~[55] &  15.2 &  13.0 &  17.1 & \multicolumn{1}{c|}{  98.1} & \ 27.5 &  13.1 &  26.0 &  99.0 & 98.55 &  18.65 & 79.90 (81.08\%) \\
\midrule
 & \multicolumn{2}{c}{CIFAR} & \multicolumn{2}{c|}{STL} & \multicolumn{2}{c}{CIFAR} & \multicolumn{2}{c|}{STL} & \multicolumn{3}{c}{/} \\ 
\midrule
CIFAR10~[56]   & \multicolumn{2}{c}{98.2} & \multicolumn{2}{c|}{ 9.8} & \multicolumn{2}{c}{99.7} & \multicolumn{2}{c|}{10.5} & 98.95 & 10.15 & 88.80 (89.74\%) \\
STL10~[57]     & \multicolumn{2}{c}{20.2} & \multicolumn{2}{c|}{98.1} & \multicolumn{2}{c}{13.6} & \multicolumn{2}{c|}{99.6} & 98.85 & 16.90 & 81.95 (82.90\%) \\         
\midrule
 & \multicolumn{2}{c}{VisDA-T} & \multicolumn{2}{c|}{VisDA-V} & \multicolumn{2}{c}{VisDA-T} & \multicolumn{2}{c|}{VisDA-V} & \multicolumn{3}{c}{/} \\ 
\midrule
VisDA-T~[59] & \multicolumn{2}{c}{99.7} & \multicolumn{2}{c|}{13.5} & \multicolumn{2}{c}{100.0} & \multicolumn{2}{c|}{10.7} & 99.85 & 12.10 & 83.07 (83.88\%) \\ 
\bottomrule
\end{tabular}
}
  \vskip -10pt
  \label{aa_3_detail}
\end{table*}

\begin{table*}[!t]
\renewcommand\arraystretch{1.6}
  \centering
    \caption{The accuracy ($\%$) of target-specified/free CUPI-Domain, CUTI-Domain~[27] and NTL~[14] on translated input data. The result of CUTI-Domain~[27] and NTL~[14] is obtained by reproducing its source code for a fair comparison. The best performance is indicated by the numbers in bold. Statistical significance (p-value $<$ 0.05~[60],~[61]) is denoted with: $^{\ast}$(CUPI-Domain vs. CUTI-Domain~[27]) and $^{\diamond}$(CUPI-Domain vs. NTL~[14]).}
  \resizebox{0.6\textwidth}{!}{
  \begin{tabular}{c|ccc|ccc}
    \toprule
    \multirow{2}{*}{Domain} & \multicolumn{3}{c|}{Target-Specified} & \multicolumn{3}{c}{Target-Free} \\
    \cline{2-7}
    & CUPI & CUTI~[27] & NTL~[14] & CUPI & CUTI~[27] & NTL~[14] \\
    \midrule   
    MNIST~[52] & 43.03  & 46.98  & 46.26 & 38.03  & 40.73  & 52.18 \\  
    USPS~[53] & 38.92  & 44.59  & 26.56 & 19.44  & 21.20  & 21.39 \\ 
    SVHN~[54] & 18.10  & 28.07  & 34.43 & 27.69  & 28.18  & 29.98 \\  
    MNIST-M~[55] & 25.90  & 42.86  & 51.60 & 15.58  & 18.13  & 13.84 \\ 
    CIFAR10~[56] & 18.76  & 23.03  & 24.11 & 16.73  & 16.52  & 17.43 \\  
    STL10~[57] & 7.82  & 8.57  & 9.40 & 8.91  & 9.71  & 9.46 \\  
    \midrule 
    Mean & \bf 12.39$^{\ast}$  & 15.45  & 15.97 & \bf 21.06$^{\ast}$  & 22.41  & 24.05 \\
    \bottomrule
  \end{tabular}}
  \label{CycleGAN}
\end{table*}

\begin{table*}[!t]
\renewcommand\arraystretch{1.6}
  \centering
    \caption{The accuracy ($\%$) of target-specified/free CUPI-Domain, CUTI-Domain~[27] and NTL~[14] on the unauthorized domain after model stealing attack by DeepSteal~[65]. The result of CUTI-Domain~[27] and NTL~[14] is obtained by reproducing its source code for a fair comparison. The best performance is indicated by the numbers in bold. Statistical significance (p-value $<$ 0.05~[60],~[61]) is denoted with: $^{\ast}$(CUPI-Domain vs. CUTI-Domain~[27]) and $^{\diamond}$(CUPI-Domain vs. NTL~[14]).}
  \resizebox{0.6\textwidth}{!}{
  \begin{tabular}{c|ccc|ccc}
    \toprule
    \multirow{2}{*}{Domain} & \multicolumn{3}{c|}{Target-Specified} & \multicolumn{3}{c}{Target-Free} \\ 
    \cline{2-7}
    & CUPI & CUTI~[27] & NTL~[14] & CUPI & CUTI~[27] & NTL~[14] \\
    \midrule
    MNIST~[52] & 7.77  & 9.24  & 22.74  & 10.11  & 11.63  & 28.60 \\ 
    USPS~[53] & 9.85  & 9.77  & 12.54  & 8.90  & 10.42  & 10.81 \\ 
    SVHN~[54] & 9.55  & 8.94  & 9.68  & 13.11  & 13.06  & 11.11 \\ 
    MNIST-M~[55] & 7.38  & 7.55  & 12.37  & 8.42  & 7.60  & 15.06 \\ 
    CIFAR10~[56] & 8.20  & 7.81  & 10.42  & 10.81  & 12.89  & 27.47 \\ 
    STL10~[57] & 9.51  & 13.41  & 13.41  & 9.77  & 12.24  & 11.98 \\
    \midrule
    Mean & \bf 8.71$^{\diamond}$  & 9.45  & 13.53  & \bf 10.19$^{\ast \diamond}$  & 11.31  & 17.51 \\
    \bottomrule
  \end{tabular}}
  \label{Steal1}
\end{table*}

\begin{table*}[!t]
\renewcommand\arraystretch{1.6}
  \centering
    \caption{The accuracy ($\%$) of target-specified/free CUPI-Domain, CUTI-Domain~[27] and NTL~[14] on the unauthorized domain after model stealing attack by Knockoff-Nets~[66]. The result of CUTI-Domain~[27] and NTL~[14] is obtained by reproducing its source code for a fair comparison. The best performance is indicated by the numbers in bold. Statistical significance (p-value $<$ 0.05~[60],~[61]) is denoted with: $^{\ast}$(CUPI-Domain vs. CUTI-Domain~[27]) and $^{\diamond}$(CUPI-Domain vs. NTL~[14]).}
  \resizebox{0.6\textwidth}{!}{
  \begin{tabular}{c|ccc|ccc}
    \toprule
    \multirow{2}{*}{Domain} & \multicolumn{3}{c|}{Target-Specified} & \multicolumn{3}{c}{Target-Free} \\ 
    \cline{2-7}
    & CUPI & CUTI~[27] & NTL~[14] & CUPI & CUTI~[27] & NTL~[14] \\
    \midrule
    MNIST~[52] & 7.42  & 20.70  & 17.45  & 10.16  & 12.76  & 12.37 \\ 
    USPS~[53] & 11.94  & 10.33  & 15.89  & 9.25  & 20.96  & 11.59 \\ 
    SVHN~[54] & 11.07  & 15.80  & 13.24  & 7.68  & 34.90  & 22.66  \\ 
    MNIST-M~[55] & 9.33  & 23.89  & 15.83  & 5.60  & 20.83  & 15.10 \\ 
    CIFAR10~[56] & 8.72  & 19.01  & 19.01  & 8.72  & 19.01  & 19.01 \\ 
    STL10~[57] & 16.02  & 11.07  & 21.75  & 16.02  & 11.07  & 21.75  \\ 
    \midrule
    Mean & \bf 10.75$^{\diamond}$  & 16.80  & 17.19  & \bf 9.57$^{\ast \diamond}$  & 19.92  & 17.08  \\
    \bottomrule
  \end{tabular}}
  \label{Steal2}
\end{table*}

\begin{table*}[!t]
\renewcommand\arraystretch{1.6}
  \centering
    \caption{The accuracy drop (relative drop) of target-specified CUPI-Domain, CUTI-Domain~[27] and NTL~[14] on Amazon Reviews Dataset~[67]. 'Authorized/Unauthorized Drop' represent the average drop (relative drop) of the method relative to SL on the authorized/unauthorized domain. The result of CUTI-Domain~[27] and NTL~[14] is obtained by reproducing its source code for a fair comparison. 
    The best performance is indicated by the numbers in bold. Statistical significance (p-value $<$ 0.05~[60],~[61]) is denoted with: $^{\ast}$(CUPI-Domain vs. CUTI-Domain~[27]) and $^{\diamond}$(CUPI-Domain vs. NTL~[14]).}
  \resizebox{1\textwidth}{!}{
  \begin{tabular}{c|ccc|ccc}
     \toprule
     \multirow{2}{*}{Domain / Drop} & \multicolumn{3}{c|}{Authorized Drop$\downarrow$} & \multicolumn{3}{c}{Unauthorized Drop$\uparrow$} \\
     \cline{2-7} 
     & CUPI-Domain & CUTI-Domain~[27] & NTL~[14] & CUPI-Domain & CUTI-Domain~[27] & NTL~[14] \\
    \midrule   
    Jewelry             & \bf 0.17 (0.24\%) & 0.40 (0.57\%) & 0.97 (1.39\%) & \bf 40.15 (66.63\%) & 35.55 (59.07\%) & 36.56 (60.60\%) \\
    Musical-Instruments & \bf 0.30 (0.48\%) & 0.50 (0.81\%) & 0.50 (0.81\%) & \bf 36.53 (60.26\%) & 35.45 (58.29\%) & 34.78 (57.11\%) \\ 
    Office-Products     & \bf 0.13 (0.21\%) & 0.67 (1.07\%) & 0.17 (0.27\%) & \bf 43.35 (71.35\%) & 42.20 (69.39\%) & 40.93 (67.15\%) \\ 
    Software            & \bf 0.07 (0.12\%) & 0.67 (1.16\%) & 0.47 (0.81\%) & \bf 49.97 (82.29\%) & 47.85 (78.67\%) & 47.51 (78.16\%) \\ 
    \midrule
    Mean                & \bf 0.17 (0.26\%)$^{\ast \diamond}$ & 0.56 (0.90\%) & 0.53 (0.82\%) & \bf 42.50 (70.13\%)$^{\ast \diamond}$ & 40.26 (66.35\%) & 39.94 (65.76\%) \\
    \bottomrule
  \end{tabular}}
  \label{text}
\end{table*}

\begin{table*}[!t]
\renewcommand\arraystretch{1.6}
  \centering
    \caption{The accuracy drop (relative drop) of target-specified CUPI-Domain, CUTI-Domain~[27] and NTL~[14] on DCASE 2018 Task 1B Dataset~[68]. 'Authorized/Unauthorized Drop' represents the average drop (relative drop) of the method relative to SL on the authorized/unauthorized domain. The result of CUTI-Domain~[27] and NTL~[14] is obtained by reproducing its source code for a fair comparison. 
    The best performance is indicated by the numbers in bold. Statistical significance (p-value $<$ 0.05~[60],~[61]) is denoted with: $^{\ast}$(CUPI-Domain vs. CUTI-Domain~[27]) and $^{\diamond}$(CUPI-Domain vs. NTL~[14]).}
  \resizebox{1\textwidth}{!}{
  \begin{tabular}{c|ccc|ccc}
     \toprule
     \multirow{2}{*}{Domain / Drop} & \multicolumn{3}{c|}{Authorized Drop$\downarrow$} & \multicolumn{3}{c}{Unauthorized Drop$\uparrow$} \\
     \cline{2-7} 
     & CUPI-Domain & CUTI-Domain~[27] & NTL~[14] & CUPI-Domain & CUTI-Domain~[27] & NTL~[14] \\
    \midrule         
    Device-A & \bf 0.00 (0.00\%) & 0.00 (0.00\%) & 2.08 (3.85\%) & \bf  6.77 (42.50\%) &  5.73 (35.00\%) & \bf 6.77 (40.00\%) \\
    Device-B & \bf 0.00 (0.00\%) & 0.00 (0.00\%) & 0.52 (1.02\%) & \bf 19.27 (71.46\%) & 16.15 (53.44\%) & 14.58 (46.96\%) \\ 
    Device-C & \bf 0.00 (0.00\%) & 0.52 (0.96\%) & \bf 0.00 (0.00\%) & \bf 23.96 (60.56\%) & 22.40 (57.44\%) & 20.83 (51.78\%) \\ 
    \midrule
    Mean     & \bf 0.00 (0.00\%)$^{\diamond}$ & 0.17 (0.32\%) & 0.87 (1.62\%) & \bf 16.67 (58.17\%)$^{\ast \diamond}$ & 14.76 (48.63\%) & 14.06 (46.25\%) \\
    \bottomrule
  \end{tabular}}
  \label{audio}
\end{table*}

\begin{table*}[!t]
\renewcommand\arraystretch{1.6}
  \centering
    \caption{The accuracy ($\%$) of target-specified CUPI-Domain on tuberculosis classification task. The vertical/horizontal axis denotes the authorized/unauthorized domain. In each task, the left of '\(\Rightarrow\)' shows the test accuracy of SL on the unauthorized domain, while the right side presents the accuracy of CUPI-Domain. 'Authorized/Unauthorized Drop' indicate the drop (relative drop) of CUPI-Domain relative to SL on authorized/unauthorized domains. The result of CUTI-Domain~[27] and NTL~[14] is obtained by reproducing its source code for a fair comparison. The best performance is indicated by the numbers in bold. Statistical significance (p-value $<$ 0.05~[60],~[61]) is denoted with: $^{\ast}$(CUPI-Domain vs. CUTI-Domain~[27]) and $^{\diamond}$(CUPI-Domain vs. NTL~[14]).}
  \resizebox{1\textwidth}{!}{
  \begin{tabular}{c|cc|ccc|ccc}
    \toprule
    \multirow{2}{*}{Domain / Drop} & \multirow{2}{*}{ChinaSet} & \multirow{2}{*}{Montgomery} & \multicolumn{3}{c|}{Authorized Drop$\downarrow$} & \multicolumn{3}{c}{Unauthorized Drop$\uparrow$} \\
    \cline{4-9}
    & & & CUPI & CUTI~[27] & NTL~[14] & CUPI & CUTI~[27] & NTL~[14] \\
    \midrule
    ChinaSet~[69]   & 89.6 $\Rightarrow$ 89.6 & 78.1 $\Rightarrow$ 15.6 & 0.00 (0.00\%) & 0.00 (0.00\%) & 0.00 (0.00\%) & 62.50 (80.03\%) & 58.30 (74.65\%) & 54.10 (69.27\%) \\ 
    Montgomery~[70] & 55.2 $\Rightarrow$ 15.6 & 90.6 $\Rightarrow$ 90.6 & 0.00 (0.00\%) & 1.00 (1.10\%) & 1.00 (1.10\%) & 39.60 (71.74\%) & 35.40 (64.13\%) & 39.60 (71.74\%) \\
    \midrule
    Mean & / & / & \bf 0.00 (0.00\%)$^{\ast \diamond}$ & 0.50 (0.55\%) & 0.50 (0.55\%) & 51.05 (75.88\%)$^{\ast \diamond}$ & 46.85 (69.39\%) & 46.85 (70.50\%) \\
    \bottomrule
  \end{tabular}}
  \label{ct_ts}
\end{table*}

\begin{table*}[!t]
\renewcommand\arraystretch{1.6}
  \centering
    \caption{The accuracy ($\%$) of target-free CUPI-Domain on tuberculosis classification task. The vertical/horizontal axis denotes the authorized/unauthorized domain. In each task, the left of '\(\Rightarrow\)' shows the test accuracy of SL on the unauthorized domain, while the right side presents the accuracy of CUPI-Domain. 'Authorized/Unauthorized Drop' indicate the drop (relative drop) of CUPI-Domain relative to SL on authorized/unauthorized domains. The result of CUTI-Domain~[27] and NTL~[14] is obtained by reproducing its source code for a fair comparison. The best performance is indicated by the numbers in bold. Statistical significance (p-value $<$ 0.05~[60],~[61]) is denoted with: $^{\ast}$(CUPI-Domain vs. CUTI-Domain~[27]) and $^{\diamond}$(CUPI-Domain vs. NTL~[14]).}
  \resizebox{1\textwidth}{!}{
  \begin{tabular}{c|cc|ccc|ccc}
    \toprule
    \multirow{2}{*}{Domain / Drop} & \multirow{2}{*}{ChinaSet} & \multirow{2}{*}{Montgomery} & \multicolumn{3}{c|}{Authorized Drop$\downarrow$} & \multicolumn{3}{c}{Unauthorized Drop$\uparrow$} \\
    \cline{4-9}
    & & & CUPI & CUTI~[27] & NTL~[14] & CUPI & CUTI~[27] & NTL~[14] \\
    \midrule
    ChinaSet~[69]   & 89.6 $\Rightarrow$ 89.6 & 78.1 $\Rightarrow$ 20.8 & 0.00 (0.00\%) & 1.10 (1.23\%) & 2.10 (2.34\%) & 57.30 (73.37\%) & 54.10 (69.27\%) & 48.90 (62.61\%) \\ 
    Montgomery~[70] & 55.2 $\Rightarrow$ 21.9 & 90.6 $\Rightarrow$ 90.6 & 0.00 (0.00\%) & 1.00 (1.10\%) & 0.00 (0.00\%) & 33.30 (60.33\%) & 30.20 (54.71\%) & 26.00 (47.10\%) \\
    \midrule
    Mean & / & / & \bf 0.00 (0.00\%)$^{\ast \diamond}$ & 1.05 (1.17\%) & 1.05 (1.17\%) & 45.30 (66.85\%)$^{\ast \diamond}$ & 42.15 (61.99\%) & 37.45 (54.86\%) \\
    \bottomrule
  \end{tabular}}
  \label{ct_tf}
\end{table*}

\end{document}